\documentclass[graybox]{svmult}

\usepackage{type1cm}        
\usepackage{makeidx}         
\usepackage{graphicx}        
\usepackage{multicol}        
\usepackage[bottom]{footmisc}
\usepackage{amsmath}
\usepackage{graphicx}
\usepackage{algorithm}
\usepackage{multirow}
\usepackage[noend]{algpseudocode}
\usepackage{float}
\usepackage{colortbl}
\definecolor{thisisgrey}{RGB}{120, 120, 120}
\usepackage{subfigure}

\usepackage{pgfplots}
\usepgfplotslibrary{colorbrewer}
\pgfplotsset{compat = 1.15, cycle list/Set1-8} 
\usetikzlibrary{pgfplots.statistics, pgfplots.colorbrewer} 
\usepackage{pgfplotstable}
\usepackage{filecontents}

\usepackage{newtxtext}       %
\usepackage[varvw]{newtxmath}       

\newcommand{\mb}[1]{\mathbf{#1}}
\newcommand{\bs}[1]{\boldsymbol{#1}}

\newcommand{\be}{\begin{equation}}
\newcommand{\ee}{\end{equation}}

\makeindex             

\begin{document}

\title*{FEMDA: a unified framework for discriminant analysis}
\author{Pierre HOUDOUIN, Matthieu JONCKHEERE, Frederic PASCAL}
\institute{Pierre HOUDOUIN \at Université Paris-Saclay, CentraleSupélec, Laboratoire des signaux et systèmes, 91190, Gif-sur-Yvette \& RTE, La Défense \email{pierre.houdouin@centralesupelec.fr}
\\
Matthieu JONCHKEERE \at LAAS, CNRS, 31400, Toulouse \email{matthieu.jonchkeere@gmail.com
\\
Frederic PASCAL \at Université Paris-Saclay, CentraleSupélec, Laboratoire des signaux et systèmes, 91190, Gif-sur-Yvette \email{frederic.pascal@centralesupelec.fr}}
}
%
%
\maketitle

\abstract{
Although linear and quadratic discriminant analysis are widely recognized classical methods, they can encounter significant challenges when dealing with non-Gaussian distributions or contaminated datasets. This is primarily due to their reliance on the Gaussian assumption, which lacks robustness. We first explain and review the classical methods to address this limitation and then present a novel approach that overcomes these issues. In this new approach, the model considered is an arbitrary Elliptically Symmetrical (ES) distribution per cluster with its own arbitrary scale parameter. This flexible model allows for potentially diverse and independent samples that may not follow identical distributions. By deriving a new decision rule, we demonstrate that maximum-likelihood parameter estimation and classification are simple, efficient, and robust compared to state-of-the-art methods.
}

\section{Introduction}
\label{sec:Intro}

In machine learning, the classification task involves assigning an observation to a group of similar elements that share homogeneous properties. During the training phase, labelled observations are used to determine all possible groups or clusters, along with their associated features. These identified features are then used to build a decision rule for classifying new observations. Classification is a research field of great interest as such problems can arise across various scientific domains. Intensive efforts have been devoted to developing diverse algorithms capable of addressing the large variety of classification problems \cite{Lotte2007Review}, \cite{Harper2005Review}, \cite{Ortiz2016Review}, \cite{Kowsari2019Text}. In this chapter, we will focus on discriminant analysis, a popular statistical-based family of classification methods.

\vspace{0.5cm}

\textbf{Discriminant analysis}

\vspace{0.5cm}

Fisher presented the first historical work on discriminant analysis in 1936 \cite{Fisher1936Use}. Fisher discriminant analysis (FDA) aims to find a linear transformation of the input data that maximizes the separation between different classes. The key idea behind the FDA is to project the original high-dimensional data onto a lower-dimensional subspace while maximizing the ratio of between-class scatter to within-class scatter. In other words, FDA seeks a projection that maximizes the separation between classes while minimizing the variation within each class.

For the two-class problem, the FDA method consists of finding the best line onto which the data should be projected to achieve linear separability. Such an optimal line is estimated using the labelled data and involves estimating the mean vector and covariance matrix for each cluster. The classification rule is then derived using the projection of the observation onto the optimal line \cite{Fisher1936Use}.

Another popular approach in discriminant analysis involves employing a discrimination rule based on estimating the likelihood that an observation belongs to a specific cluster. In this framework, we assume that each observation is generated according to a probability density function $p_k$ that only depends on the cluster $\mathcal{C}_k$ it belongs to. We typically suppose that all $p_k$ belong to a parametric family of distributions: $\forall k, p_k \in \{ p_{\bs{\theta}}, \bs{\theta} \in \bs{\Theta} \}$. Discriminant analysis methods then boil down to two steps:
\begin{itemize}
    \item The first one is a supervised learning problem. It consists of finding the optimal parameter ${\hat{\bs{\theta}}}_k \in \bs{\Theta}$ that describes best each cluster using the labelled data of each cluster. 
    \item The second step is the derivation of the decision rule to classify the unlabelled observations. The goal is to choose the cluster $\mathcal{C}_k$ that maximizes the likelihood $p_k(\mb{x})$ conditioned by the cluster.
\end{itemize}

In the first works that employed a discrimination rule based on likelihood estimation, it was assumed that all the $p_k$ followed a multivariate Gaussian distribution. As discussed by Huberty in \cite{Huberty1975Discriminant} in its review of the different discriminant analysis techniques, the supervised learning part involves using the labelled data to estimate each cluster's mean and covariance matrix. The decision rule for classifying unlabelled observations then boils down to maximizing the likelihood, computed using the estimated parameters and the multivariate Gaussian probability density function. When we introduce the assumption of homoscedasticity (i.e. same covariance matrix for all clusters), the clusters become linearly separable, and the resulting decision rule is close to the one obtained by Fisher \cite{Fisher1936Use}. It is referred to as Linear Discriminant Analysis (LDA), where a linear combination of the features is used to predict the label of an observation. If the homoscedasticity assumption, however, does not hold, the clusters are no longer linearly separable, and the resulting decision rule is referred to as Quadratic Discriminant Analysis (QDA), where classification is performed by considering quadratic combinations of the features \cite{Tharwat2016Linear}.

\vspace{0.5cm}

\textbf{Robust discriminant analysis}

\vspace{0.5cm}

One of the main drawbacks of traditional discriminant analysis methods is their sensitivity to certain assumptions. When one underlying assumption fails to hold, it can significantly impact the results. This issue was already studied in 1966 by Lachenbruch \cite{Lachenbruch1966Discriminant}, who investigated how misclassified samples in the training set affect the performances. In 1979, together with Goldstein \cite{Goldstein1979Discriminant}, they examined the influence of data contamination on the maximum likelihood approach and provided formulas for the error rates. In the same year, Lachenbruch \cite{Lachenbruch1979How} also published an article on how the non-normality assumption affects QDA, concluding that misclassification rates increase with high skewness and, to a smaller extent, high kurtosis. In answer to this sensitivity, robust methods were developed to improve robustness. To alleviate the sensitivity of mean vectors and covariance matrix estimation in FDA, Kim \cite{Kim2006Robust} proposed to explicitly incorporate a model of data uncertainty and optimize the worst-case scenario. This model, however, is only suitable for binary classification. In 1978, Randles \cite{Randles1978Generalized} suggested improving the robustness of the FDA by assigning weights to observations based on their proximity to overlapping regions, again in the case of binary classification. The goal was to reduce the influence of observations that are far from the average of mean vectors, thereby mitigating the impact of outliers. The author also provides LDA and QDA approaches using robust M-estimators, developed by Huber in 1977 \cite{Huber1977Robust}, to reduce the sensitivity of mean vectors and covariance matrix estimation to outliers. The main limitation of M-estimators is their low breakpoint (proportion of outliers estimator can handle) in high dimension, as highlighted by Maronna \cite{Maronna1976Robust}. This led to recent research focusing on higher breakpoint options, such as Chork \& Rousseuw in 1992 \cite{Chork1992Integrating} with the minimum volume ellipsoid estimator, Hawkins \& McLachlan in 1997 \cite{Hawkins1997High} with the minimum within-group covariance determinant and more recently Croux \& Dehon in 2001 \cite{Croux2001Robust} that examined the performances of discriminant analysis with high breakpoint S-estimators. In 2004, Hubert \& Van Driessen \cite{Hubert2004Fast} studied the performances of the minimum covariance discriminant estimator, which offers robustness and computational efficiency, making it suitable for large datasets. Finally, in 2019, Wen \& Fang \cite{Wen2019Robust} proposed an extension of FDA, called robust sparse linear discriminant analysis, which incorporates feature selection along with the projection.

\vspace{0.5cm}

\textbf{Generalized discriminant analysis}

\vspace{0.5cm}

Many studies have been conducted to address the limitations associated with the Gaussian underlying model assumption and reduce its sensitivity. In 1992, Wakaki \cite{Wakaki1994Discriminant} investigated the expression of Fisher's linear discriminant function for the very large family of elliptical symmetric distributions with a common covariance matrix, aiming to extend the applicability of discriminant analysis beyond the Gaussian distributions. Different approaches such as marginal Fisher analysis \cite{Huang2019Multiple}, discriminative locality alignment \cite{Zhang2008Discriminative}, and manifold partition discriminant analysis \cite{Zhou2017Manifold} were also developed. These three methods use both nearby observations and label information to come up with a more general projection that can deal more efficiently with nonlinear structures. Regarding likelihood-based approaches, namely LDA and QDA, Andrews et al. \cite{Jeffrey2011Model} examined in 2010 the classification with a mixture of multivariate-t distributions. They developed an algorithm for parameter estimation across four distinct scenarios, which involved considering equality or inequality of covariance matrices and variations in degrees of freedom across different clusters. By considering more flexible distributions, these approaches aimed to enhance the robustness and performance of discriminant analysis methods.

Bose et al. \cite{Bose2015Generalized} in 2015 attempted to generalize LDA and QDA to elliptically symmetric distributions. The proposed method is called Generalized Quadratic Discriminant Analysis (GQDA) and is semi-parametric. It involves estimating a threshold parameter $c$ whose optimal value is fixed for all sub-families of distributions. For example, when  $c=1$, GQDA corresponds to the classic QDA for a mixture of multivariate Gaussian distributions, while $c=0$ boils down to LDA. In the case of multivariate-t distributions, the optimal value of $c$ is determined by the degrees of freedom of each cluster's distribution. Of course, in practical applications, the sub-family is unknown, necessitating the estimation of the threshold parameter. Finally, Ghosh et al. \cite{Ghosh2020Robust} in 2020 used Bose's work to handle situations where the sensitive Gaussianity hypothesis was not fulfilled. They also explored using different robust estimators for the mean and covariance matrix to handle outliers. Their work resulted in Robust Generalized Quadratic Discriminant Analysis (RGQDA), a new robust discriminant analysis method.

With the rise of neural networks, novel approaches to discriminant analysis have been developed. In 2016, Dorfer et al. \cite{Dorfer2016Deep} introduced Deep LDA, a new deep neural network method. Deep LDA can be considered as a non-linear extension of LDA, aiming to minimize intra-cluster variance while maximizing extra-cluster variance, able to capture complex relationships and patterns in the data. A last approach that deserves to be highlighted is the study of distribution-free discriminant analysis by Burr and Doak in 1996 \cite{Burr1999Distribution}. This approach uses the kernel method developed by Sivlerman in 1986 \cite{Sivlerman1986Density} to estimate the probability density function of each cluster. Although the approach is radically different from the one proposed in this article, the goal is the same: build a decision rule for discriminant analysis without relying on prior assumptions about the distributions and handle clusters with different distributions.

\vspace{0.5cm}

\textbf{Our contribution}

\vspace{0.5cm}

This work introduces a novel, robust, and model-free discriminant analysis algorithm. Our method does not assume that points within the same cluster are drawn from the same distribution. Instead, we allow each observation to be drawn from its own elliptically symmetric distribution; see Chapter 1, \cite{Boente2014Characterization} and \cite{Ollila2012Complex} for more information on this family of distributions. This provides greater flexibility in the modelling as we only assume that observations are independent.

Since each point can have its own distribution, observations inside the same cluster are not necessarily identically distributed. We are no longer in the framework where each $\mb{x} \in \mathcal{C}_k$ is drawn from $p_k$; in this framework, each $\mb{x}_i$ is drawn from $p_{i,k}$ \cite{Roizman2020Flexible}. The conditions to belong to the same cluster are consequently redefined. The precise definition is provided in the next section.

The counterpart of such flexibility in the modelling does reverberate in the more permissive clusters' membership conditions. Observations within the same cluster only fulfil proportional covariance matrices: their covariance matrix share the same direction, but the scale may differ. While this may appear insufficient for describing a cluster in low-dimensional cases, it remains very satisfying in higher dimensions. This increased flexibility also poses challenges for the likelihood estimation, as there are many more unknown parameters to consider. To deal with this issue, we will look to maximize the marginal likelihood, which is an average of the likelihood computed over the scale parameter of the covariance matrix. The structure of the paper is as follows: in Section 2, we present the model along with the theoretical framework, providing the main derivations of this work. Section 3 encompasses the experiments conducted on synthetic data. Section 4 contains the real data experiments, and conclusions and perspectives are drawn in section 5.

\section{Model and theoretical contribution}
\label{sec:Hilbert}

\subsection{Presentation of the statistical modelling}

Let $\mb{x}_1,...,\mb{x}_n \in \mathbb{R}^m$ be $n$ \textbf{independent} observations splitted into $K$ different classes. Each cluster $\mathcal{C}_k$ comprises $n_k$ observations. Previous works in discriminant analysis usually assume that points in the same cluster are drawn from the same distribution. In our contribution, we discharge this hypothesis. We allow each observation $\mb{x}_i$ to:

\begin{itemize}
    \item Be drawn from \textbf{its own elliptically symmetric distribution} with density generator function $g_{i,k}$,
    \item Have its own covariance matrix, written as $\bs{\Sigma}_{i,k} = \tau_{i,k} \bs{\Sigma}_k$, $\tau_{i,k}$ being an unknown scale factor of the observation and $\bs{\Sigma}_k$ the common scatter matrix of points of cluster $\mathcal{C}_k$,
    \item Have a prior probability density function $h_{i,k}$ on the scale factor.
\end{itemize}

In this new modelling, the joint probability density function is

\begin{align*}
f(\mb{x}_i \in \mathcal{C}_k, \tau_{i,k}) =& \frac{\Gamma(\frac{m}{2})h_{i,k}(\tau_{i,k})}{\pi^{\frac{m}{2}}\int_0^{+ \infty}t^{\frac{m}{2}-1}g_{i,k}(t)dt}|\bs{\Sigma}_k|^{-\frac{1}{2}} \tau_{i,k}^{-\frac{m}{2}} g_{i,k}\left( \frac{(\mb{x}-\bs{\mu}_k)^T \bs{\Sigma}_k^{-1} (\mb{x}-\bs{\mu}_k)}{\tau_{i,k}} \right) \\
\mb{x}_i \in \mathcal{C}_k \sim& \bs{\mu}_k + \sqrt{\tau_{i,k}} \sqrt{Q_{i,k}} \bs{\Sigma}_k^{\frac{1}{2}} \mb{u_i}.
\end{align*}

The second equation is the stochastic representation of $\mb{x}_i$.  The vector $\mb{u}_i$ is drawn using the uniform distribution over the $m$-dimensional unit sphere while $Q_{i,k}$ is a positive random variable directly linked to $g_{i,k}$ by the relation: $p_{Q_{i,k}}(t) = \frac{t^{\frac{m}{2}-1}g_{i,k}(t)}{\int_0^{\infty}t^{\frac{m}{2}-1}g_{i,k}(t)dt}$. The double dependence regarding $i$ and $k$ for a parameter $\bs{\theta} \in \bs{\Theta}$ implies that each point of the dataset has its own value $\bs{\theta}_{i,k}$, belonging to a subset of $\bs{\Theta}$ depending on $k$: $\bs{\theta}_{i,k} \in \bs{\Theta}_k \subseteq \bs{\Theta}$. Since both the cluster and the observation influence the joint probability density function $p_{i,k}$, belonging to the same cluster can no longer be defined as sharing the same probability density function. Belonging to cluster $\mathcal{C}_k$ is now characterized as follows:

\begin{itemize}
    \item Respecting the constraints on the density generator function $g_{i,k}$ and the prior $h_{i,k}$,
    \item Have the mean vector equal to $\bs{\mu}_k$,
    \item Have the scatter matrix equal to $\bs{\Sigma}_k$.
\end{itemize}

This new modelling is very flexible and provides a high level of generality. Observations are not i.i.d anymore, but only independent. The family of distributions used is significantly larger than the traditional multivariate Gaussian or student distribution. The density generator function $g_{i,k}$ can be any function $\mathbb{R}^+ \longrightarrow \mathbb{R}^+$ that fulfills the condition $\int_{\mathbb{R}^+}t^{\frac{m}{2}}g_{i,k}(t)dt < +\infty$ and $h_{i,k}$ is any prior defined on $\mathbb{R}^{+*}$. In this framework, the constraints on the density generator and the prior are fixed by the user, whereas $\bs{\mu}_k$ and $\bs{\Sigma}_k$ are unknown parameters that must be estimated. Each choice of $(g_{i,k}, h_{i,k})$ leads to a different discriminant analysis method. Besides, all the usual discriminant analysis methods (QDA, t-QDA...) can be retrieved with the appropriate choice on the density generator and the prior. Details for many famous methods are provided later on. To obtain our classification algorithm, we aim to stick as long as possible to the general case of unknown $(g_{i,k}, h_{i,k})$. In the next subsection, we are first going to derive the estimators of $\bs{\mu}_k$ and $\bs{\Sigma}_k$ and the decision rule in this general case, and then add an assumption on $h_{i,k}$ in order to obtain a tractable algorithm.

\subsection{General case estimators and decision rule}

To obtain estimators for the unknown scatter and location parameters, one has to efficiently deal with the last unknown parameter, $\tau_{i,k}$. Two options are possible:
\begin{itemize}
    \item Use the completed likelihood, the likelihood where $\tau_{i,k}$ is replaced by its maximum likelihood estimator,
    \item Use a Bayesian approach with the maximum marginal likelihood estimators.
\end{itemize}

The authors have already applied the first approach \cite{HOUDOUIN2022Robust} in a less general case. In this previous work, there are more stringent hypotheses (the density generator function only depended on $i$, and no prior was assumed on $\tau_{i,k}$) and used the completed likelihood to come up with the Flexible EM-inspired Discriminant Analysis (FEMDA) algorithm. In this paper, we will focus on the second Bayesian approach. The marginal likelihood of $\mb{x}_1,...,\mb{x}_{n_k} \in \mathcal{C}_k$ relatively to $\bs{\mu}_k$ and $\bs{\Sigma}_k$ is

\begin{align*}
    \mathcal{L}(\mb{x};\bs{\mu}_k, \bs{\Sigma}_k)= \int_{]0,+\infty[^{n_k}} \prod_{i=1}^{n_k} p_{i,k}\left(\mb{x}_i, \tau_{i,k};\bs{\mu}_k, \bs{\Sigma}_k\right) d\tau_{i,k}.
\end{align*}

\textbf{Proposition 1.1}

\vspace{0.4cm}

Assuming that $h_{i,k}(t)$ is an integrable function with respect to the Lebesgue measure, the marginal likelihood of the sample $\mb{x}_k = \left( \mb{x}_1,...\mb{x}_{n_k} \right)$ belonging to cluster $\mathcal{C}_k$ relatively to $\bs{\mu}_k$ and $\bs{\Sigma}_k$ in this theoretical framework is
\begin{align*}
    \mathcal{L} = \prod_{i=1}^{n_k} &\frac{\Gamma(\frac{m}{2}) |\bs{\Sigma}_k|^{-\frac{1}{2}} \left( (\mb{x}_i-\bs{\mu}_k)^T \bs{\Sigma}_k^{-1}(\mb{x}_i-\bs{\mu}_k)\right)^{1-\frac{m}{2}}}{\pi^{\frac{m}{2}} \int_0^{+ \infty}t^{\frac{m}{2}-1}g_{i,k}(t)dt} \\
    &  \int_0^{+\infty} h_{i,k}\left(\frac{(\mb{x}_i-\bs{\mu}_k)^T \bs{\Sigma}_k^{-1} (\mb{x}_i-\bs{\mu}_k)}{t}\right) t^{\frac{m}{2}-2}g_{i,k}(t) dt. 
\end{align*}

\textit{Proof}: See Section 6

\vspace{0.4cm}

\textbf{Remark 1.2}

\vspace{0.4cm}

Choosing the Dirac measure $h_{i,k}(\tau_{i,k})d\tau = \delta_1(\tau_{i,k})$ allows us to come back to the traditional discriminant analysis modelling. In this particular case, the framework, however, changes since $h_{i,k}(\tau_{i,k})$ is no longer a probability density function defined over the Lebesgue measure. In this particular case, the use of the Maximal Marginal Likelihood Estimator (MMLE) boils down to the classic Maximum Likelihood Estimator (MLE). We will now derive the analytical expression of the MMLE by maximizing the marginal likelihood of the sample with respect to $\bs{\mu}_k$ and $\bs{\Sigma}_k$. Once the estimators are obtained, we can obtain the expression of the decision rule for a new observation $\mb{x}$. It consists in choosing the cluster that will maximize its marginal likelihood $\mathcal{L}(\mb{x}|\mb{x} \in \mathcal{C}_k)$. According to Neyman-Pearson's lemma, as distributions here are not degenerate, the unique Uniformly Most Powerful (UMP) statistical test is used to classify the observation \cite{Wald1942Chapter}.

\vspace{0.4cm}

\textbf{Theorem 2.1}

\vspace{0.4cm}

Assuming that $h_{i,k}$ is differentiable, the expressions of the MMLE estimators of the location and scale parameters, as well as the decision rule, are
\begin{align*}
    \hat{\bs{\mu}}_k =& \frac{\sum_{i=1}^{n_k} w_{i,k} \mb{x}_i}{\sum_{i=1}^{n_k} w_{i,k}} \\
    \hat{\bs{\Sigma}}_k =& \frac{m-2}{n_k} \sum_{i=1}^{n_k} w_{i,k} (\mb{x}_i-\hat{\bs{\mu}}_k)(\mb{x}_i-\hat{\bs{\mu}}_k)^T \\
    w_{i,k} =& \frac{1}{(\mb{x}_i-\hat{\bs{\mu}}_k)^T\hat{\bs{\Sigma}}_k^{-1}(\mb{x}_i-\hat{\bs{\mu}}_k)} \\
    &-\frac{2}{m-2}\frac{\int_0^{+\infty}h_{i,k}'\left(\frac{(\mb{x}_i-\hat{\bs{\mu}}_k)^T\hat{\bs{\Sigma}}_k^{-1}(\mb{x}_i-\hat{\bs{\mu}}_k)}{t}\right)t^{\frac{m}{2}-3}g_{i,k}(t)dt}{\int_0^{+\infty}h_{i,k}\left(\frac{(\mb{x}_i-\hat{\bs{\mu}}_k)^T\hat{\bs{\Sigma}}_k^{-1}(\mb{x}_i-\hat{\bs{\mu}}_k)}{t}\right)t^{\frac{m}{2}-2}g_{i,k}(t)dt} \\
     k =& \arg \max_{k \in [1,K]} \frac{|\hat{\bs{\Sigma}}_k|^{-\frac{1}{2}} \left( (\mb{x}_i-\hat{\bs{\mu}}_k)^T \hat{\bs{\Sigma}}_k^{-1}(\mb{x}_i-\hat{\bs{\mu}}_k)\right)^{1-\frac{m}{2}}}{\int_0^{+ \infty}t^{\frac{m}{2}-1}g_{i,k}(t)dt} \\
     &\int_0^{+\infty} h_{i,k}\left(\frac{(\mb{x}_i-\hat{\bs{\mu}}_k)^T \hat{\bs{\Sigma}}_k^{-1} (\mb{x}_i-\hat{\bs{\mu}}_k)}{t}\right) t^{\frac{m}{2}-2}g_{i,k}(t) dt.
\end{align*}

\textit{Proof}: See Section 7

\vspace{0.4cm}

\textbf{Remark 2.2}

\vspace{0.4cm}

At this stage, let us underline the specific parametric case of $h_{i,k}=h_k=h_{\alpha_k}(.)$ and $g_{i,k}=g$ where $\alpha_k$ is a class parameter, which will be used later on to deal with the cases of compound Gaussian distributions. In this situation, the expression of the estimators of $\bs{\mu}_k$ and $\bs{\Sigma}_k$ becomes tractable if we can estimate $\alpha_k$. This can be done by solving the following optimization problem. 

\begin{align*}
    \hat{\alpha}_k &= \arg \max_{\alpha_k \in \bs{\Theta}} \sum_{i=1}^{n_k} \log \left( \int_0^{\infty} h_{\alpha_k}\left(\frac{(\mb{x}_i-\hat{\bs{\mu}}_k)^T\hat{\bs{\Sigma}}_k^{-1}(\mb{x}_i-\hat{\bs{\mu}}_k)}{t}\right)t^{\frac{m}{2}-2}g(t)dt \right).
\end{align*}

At this point, both the estimators and the decision rule are intractable in the general case. We will add assumptions on the prior distribution of $\tau_{i,k}$ to obtain our algorithm.

\subsection{Derivation of the algorithm}

\subsubsection{Non-informative prior $h_{i,k}$}
 
If we do not have specific information on the nuisance parameters $\tau_{i,k}$, then using the non-informative Jeffrey's prior \cite{jeffreys1946invariant}. In the particular case of a uni-dimensional parameter, Jeffrey's prior, which may be an improper prior, is proportional to the square root of the Fisher information. Based on \cite{besson2013fisher}, we obtain the following result.

\vspace{0.4cm}

\textbf{Proposition 3}

\vspace{0.4cm}

The expression of the non-informative Jeffrey's prior on $\tau_{i,k}$ is given by 

\begin{align*}
     h_{i,k}(t) \propto \frac{1}{t}.
\end{align*}

\textit{Proof}: By definition, the expression of Jeffrey's prior for a one-dimensional parameter $\tau$ is 

\begin{align*}
    & h_{i,k}(\tau) \propto \sqrt{\mathcal{I}_{i,k}(\tau)}, \mbox{with } \mathcal{I}_{i,k}(\tau) =-\mathbb{E}_{\mb{X_i}} \left[ \frac{\partial^2\left(\log \circ p_{i,k}(\mb{x_i},\tau)\right)}{\partial \tau^2} \right] \\
    & \mbox{ being the Fisher information.}
\end{align*}

We define $d_{i,k}=(\mb{x_i}-\bs{\mu}_k)^T \bs{\Sigma}_k^{-1} (\mb{x_i}-\bs{\mu}_k)$ and $q_{i,k} = \frac{(\mb{x_i}-\bs{\mu}_k)^T \bs{\Sigma}_k^{-1} (\mb{x_i}-\bs{\mu}_k)}{\tau}$. Using the stochastic form, one can write $\forall \mb{x_i} \in \mathbb{R}^m$

\begin{align*}
    &\mb{x_i} \sim \bs{\mu}_k + \sqrt{\tau} \sqrt{Q_{i,k}} \bs{\Sigma}_k^{\frac{1}{2}} \bs{u_i} \\
    \iff &\tau^{-\frac{1}{2}} \bs{\Sigma}_k^{-\frac{1}{2}}(\mb{x_i}-\bs{\mu}_k) \sim \sqrt{Q_{i,k}} \bs{u_i} \\
   \iff &q_{i,k} \sim Q_{i,k} \mbox{ since } \bs{u_i}^T \bs{u_i} = 1.
\end{align*}

We can now compute the Fisher information:

\begin{align*}
    \mathcal{I}_{i,k}(\tau) =& \mathbb{E}_{\mb{X_i}} \left[ \left(\frac{\partial\left(\log \circ p_{i,k}(\mb{x_i},\tau)\right)}{\partial \tau} \right)^2 \right] =-\mathbb{E}_{\mb{X_i}} \left[ \frac{\partial^2\left(\log \circ p_{i,k}(\mb{x_i},\tau)\right)}{\partial \tau^2} \right] \\
    =&-\mathbb{E}_{Q_{i,k}} \left[ \frac{m}{2\tau^2} + \frac{2}{\tau^2} \frac{q_{i,k} g_{i,k}'(q_{i,k})}{g_{i,k}(q_k)} + \frac{1}{\tau^2} q_{i,k}^2 \frac{g_{i,k}''(q_{i,k})g_{i,k}(q_{i,k})-g_{i,k}'(q_{i,k})^2}{g_{i,k}(q_{i,k})^2} \right] \\
    =& \frac{-m}{2\tau^2} - \frac{2}{\tau^2} \int_0^{\infty} \frac{qg_{i,k}'(q)}{g_{i,k}(q)} \frac{q^{\frac{m}{2}-1}g_{i,k}(q)}{\int_0^{\infty}t^{\frac{m}{2}-1}g_{i,k}(t)dt}dq \\
    &- \frac{1}{\tau^2} \int_0^{\infty} q^2 \frac{g_{i,k}''(q)g_{i,k}(q)-g_{i,k}'(q)^2}{g_{i,k}(q)^2} \frac{q^{\frac{m}{2}-1}g_{i,k}(q)}{\int_0^{\infty}t^{\frac{m}{2}-1}g_{i,k}(t)dt}dq
\end{align*}

\begin{align*}
    =& \frac{-m}{2\tau^2} - \frac{2}{\tau^2} \int_0^{\infty} \frac{q^{\frac{m}{2}}g_{i,k}'(q)}{\int_0^{\infty}t^{\frac{m}{2}-1}g_{i,k}(t)dt}dq \\
    &- \frac{1}{\tau^2} \int_0^{\infty} q^{\frac{m}{2}+1} \frac{g_{i,k}''(q)g_{i,k}(q)-g_{i,k}'(q)^2}{g_{i,k}(q)\int_0^{\infty}t^{\frac{m}{2}-1}g_{i,k}(t)dt}dq \\
    =& \frac{-m}{2\tau^2} + \frac{m}{\tau^2} \int_0^{\infty} \frac{q^{\frac{m}{2}-1}g_{i,k}(q)}{\int_0^{\infty}t^{\frac{m}{2}-1}g_{i,k}(t)dt}dq \\
    &-\frac{1}{\tau^2} \int_0^{\infty} q^{\frac{m}{2}+1} \frac{g_{i,k}''(q)g_{i,k}(q)-g_{i,k}'(q)^2}{g_{i,k}(q)\int_0^{\infty}t^{\frac{m}{2}-1}g_{i,k}(t)dt}dq \\
    =&\frac{1}{\tau^2} \left(\frac{m}{2} - \int_0^{\infty} q^{\frac{m}{2}+1} \frac{g_{i,k}''(q)g_{i,k}(q)-g_{i,k}'(q)^2}{g_{i,k}(q)\int_0^{\infty}t^{\frac{m}{2}-1}g_{i,k}(t)dt}dq \right).
\end{align*}

Thus, we do indeed have 

\begin{align*}
    h_{i,k}(t) \propto \frac{\sqrt{\frac{m}{2} - \int_0^{\infty} q^{\frac{m}{2}+1} \frac{g_{i,k}''(q)g_{i,k}(q)-g_{i,k}'(q)^2}{g_{i,k}(q)\int_0^{\infty}t^{\frac{m}{2}-1}g_{i,k}(t)dt}dq}}{t}.
\end{align*}

Let us make a few remarks:
\begin{itemize}
    \item Since $t \longrightarrow t^{\frac{m}{2}-1}g_{i,k}(t)$ is integrable on $\mathbb{R}^+$, then $lim_{\infty} t^{\frac{m}{2}}g_{i,k}(t) = 0$, which justifies the integration by part to reach the last line,
    \item The Fisher information is always positive since it can also be written as the expected value of a positive random variable, so the square root is always well-defined.
\end{itemize}

\subsubsection{Decision rule and estimators under Jeffrey's prior}

We can now derive the expression of the estimators as well as the decision rule using Jeffrey's prior. Since $h(t) \propto \frac{C}{t}$ is differentiable, Theorem 2.1 applies.

\vspace{0.4cm}

\textbf{Theorem 4}

\vspace{0.4cm}

If we suppose that we have the non-informative Jeffrey's prior on the $\tau_{i,k}$, the decision rule and the estimators of location and scale parameters are

\begin{align*}
    \hat{\bs{\mu}}_k &= \frac{\sum_{i=1}^{n_k} w_{i,k} \mb{x}_i}{\sum_{i=1}^{n_k} w_{i,k}} \\
    \hat{\bs{\Sigma}}_k &= \frac{m}{n_k} \sum_{i=1}^{n_k} w_{i,k} (\mb{x}_i-\hat{\bs{\mu}}_k)(\mb{x}_i-\hat{\bs{\mu}}_k)^T \\
    w_{i,k} &= \frac{1}{(\mb{x}_i-\hat{\bs{\mu}}_k)^T\hat{\bs{\Sigma}}_k^{-1}(\mb{x}_i-\hat{\bs{\mu}}_k)} \\
    k &= \arg \min_{k \in [1,K]} \log \left( (\mb{x}_i-\hat{\bs{\mu}}_k)^T \hat{\bs{\Sigma}}_k^{-1}(\mb{x}_i-\hat{\bs{\mu}}_k) \right) + \frac{1}{m} \log(|\hat{\bs{\Sigma}}_k|).
\end{align*}

\textit{Proof}: See Section 8

\vspace{0.4cm}

At this stage, one can notice that the resulting estimators are M-estimators that are intrinsically robust despite being derived from an MMLE approach. They can be obtained using $\rho(\mb{x}, \bs{\mu}, \bs{\Sigma})=\log(|\bs{\Sigma}|) + \log \left( (\mb{x}-\bs{\mu})^T \bs{\Sigma}^{-1} (\mb{x}-\bs{\mu}) \right)$. Both estimators are also very close to Tyler's M-estimators. The only difference arises in the estimator of the location parameter, as Tyler obtains a square root at the weight's denominator. Another interesting remark is that first, $\hat{\bs{\mu}}_k$ is insensitive to the scale of $\hat{\bs{\Sigma}}_k$ and if $\left(\bs{\mu}, \bs{\Sigma}\right)$ is a solution to this coupled fixed-point equation, $\left(\bs{\mu}, \lambda \bs{\Sigma}\right)$ is also a solution. The same remark can be drawn for the decision rule: replacing $\bs{\Sigma}$ by $\lambda \bs{\Sigma}$ does not change the classification. To conclude, both our estimators and decision rule are scale-insensitive. Although the approach is different, using the MMLE instead of the completed likelihood, we achieve the same estimators as in \cite{HOUDOUIN2022Robust}. 

\subsection{Retrieval of other usual discriminant analysis methods}

Jeffrey's prior choice leads to the FEMDA algorithm's derivation. However, different choices on $g_{i,k}$ and $h_{i,k}$ allow the retrieval of many other well-known discriminant analysis methods.

\subsubsection{Quadratic Discriminant Analysis}

QDA assumes that the points of each cluster are drawn from a multivariate Gaussian distribution with parameters $\bs{\mu}_k$ and $\bs{\Sigma}_k$. The MLE of the parameters and the decision rule are

\begin{align*}
    \hat{\bs{\mu}}_k &= \frac{1}{n_k} \sum_{i=1}^{n_k} \mb{x}_i \\
    \hat{\bs{\Sigma}}_k &= \frac{1}{n_k} \sum_{i=1}^{n_k} (\mb{x}_i-\hat{\bs{\mu}}_k)(\mb{x}_i-\hat{\bs{\mu}}_k)^T \\
    k &= \arg \min_{k \in [1,K]} (\mb{x}_i-\hat{\bs{\mu}}_k)^T \hat{\bs{\Sigma}}_k^{-1}(\mb{x}_i-\hat{\bs{\mu}}_k).
\end{align*}

One can write

\begin{align*}
    &\mb{x}_i \in \mathcal{C}_k \sim \bs{\mu}_k + \bs{\Sigma}_k^{\frac{1}{2}} \mathcal{N}(0, I) \\
    p(&\mb{x}_i \in \mathcal{C}_k) = (2\pi)^{-\frac{m}{2}} |\bs{\Sigma}_k|^{-\frac{1}{2}} e^{-\frac{1}{2}(\mb{x}_i-\bs{\mu}_k)^T\bs{\Sigma}_k^{-1}(\mb{x}_i-\bs{\mu}_k)}.
\end{align*}

This corresponds to the choices:
\begin{itemize}
    \item $h_{i,k}(\tau) d\tau = \delta_1(\tau)$,
    \item $g_{i,k}(t) = g(t) =e^{-\frac{\mb{t}}{2}}$.
\end{itemize}

Here, the prior $h$ is no longer defined over the Lebesgue measure but over the Dirac measure. Thus, neither Proposition 1.1 nor Theorem 2.1 hold. 

\subsubsection{$t$-QDA}

$t$-QDA assumes that the points of each cluster are drawn from a multivariate student distribution with parameters $\nu_k$, $\bs{\mu}_k$ and $\bs{\Sigma}_k$. The MLE of the parameters are

\begin{align*}
    \hat{\bs{\mu}}_k =& \frac{\sum_{i=1}^{n_k}w_{i,k}\mb{x}_i}{\sum_{i=1}^{n_k}w_{i,k}} \\
    \hat{\bs{\Sigma}}_k =& \frac{\hat{\nu}_k + m}{n_k}\sum_{i=1}^{n_k} w_{i,k} (\mb{x}_i-\hat{\bs{\mu}}_k)(\mb{x}_i-\hat{\bs{\mu}}_k)^T \\
    \hat{\nu}_k =& \arg \min_{\nu \in \mathbb{R}^+} \frac{m}{2}\log(\nu) + \log \Gamma\left( \frac{\nu}{2} \right) - \log \Gamma\left( \frac{\nu + m}{2} \right) \\ 
    &+ \frac{\nu + m}{2n_k} \sum_{i=1}^{n_k} \log\left( 1 + \frac{(\mb{x}_i-\hat{\bs{\mu}}_k)^T \hat{\bs{\Sigma}}_k^{-1}(\mb{x}_i-\hat{\bs{\mu}}_k)}{\nu} \right) \\
    w_{i,k} =& \frac{1}{\hat{\nu}_k + (\mb{x}_i-\hat{\bs{\mu}}_k)^T \hat{\bs{\Sigma}}_k^{-1}(\mb{x}_i-\hat{\bs{\mu}}_k)}.
\end{align*}

Multivariate student distribution belongs to the family of compound Gaussian distributions. Thus, one can write:

\begin{align*}
    &\mb{x}_i \in \mathcal{C}_k \sim \bs{\mu}_k + \sqrt{\frac{1}{\Gamma(\frac{\nu_k}{2}, \frac{2}{\nu_k})}} \bs{\Sigma}_k^{\frac{1}{2}} \mathcal{N}(0, I) \\
    p(&\mb{x}_i \in \mathcal{C}_k) = \frac{\Gamma(\frac{\nu_k+m}{2})}{\Gamma(\frac{\nu_k}{2})\nu_k^{\frac{m}{2}}\pi^{\frac{m}{2}}} |\bs{\Sigma}_k|^{-\frac{1}{2}} \left( 1 + \frac{(\mb{x}_i-\bs{\mu}_k)^T\bs{\Sigma}_k^{-1}(\mb{x}_i-\bs{\mu}_k)}{\nu_k} \right)^{-\frac{\nu_k+m}{2}}.
\end{align*}

This corresponds to the choices:
\begin{itemize}
    \item $h_{i,k}(\tau) = \frac{\left(\frac{\nu_k}{2}\right)^{\frac{\nu_k}{2}}e^{-\frac{\nu_k}{2 \tau}}}{\Gamma\left(\frac{\nu_k}{2}\right)\tau^{1+\frac{\nu_k}{2}}}$,
    \item $g_{i,k}(t) = g(t) = e^{-\frac{\mb{x}}{2}}$.
\end{itemize}

Here, $h_{i,k}$ is differentiable and has the form of $h_{\alpha_k}(.)$ so both Theorem 2.1 and Remark 2.2 apply. It does allow us to retrieve the estimators of the degree of freedom, scale, and location parameters, as well as the decision rule:

\begin{align*}
    \hat{\bs{\mu}}_k =& \frac{\sum_{i=1}^{n_k} w_{i,k} \mb{x}_i}{\sum_{i=1}^{n_k} w_{i,k}} \\
    \hat{\bs{\Sigma}}_k =& \frac{m-2}{n_k} \sum_{i=1}^{n_k} w_{i,k} (\mb{x}_i-\hat{\bs{\mu}}_k)(\mb{x}_i-\hat{\bs{\mu}}_k)^T \\
    w_{i,k} =& \frac{\nu_k+m}{(m-2)\left(\nu_k+(\mb{x}_i-\hat{\bs{\mu}}_k)^T\hat{\bs{\Sigma}}_k^{-1}(\mb{x}_i-\hat{\bs{\mu}}_k)\right)} \\
    \hat{\nu}_k =& \arg \min_{\nu \in \mathbb{R}^+} \frac{m}{2}\log(\nu) + \log \Gamma\left( \frac{\nu}{2} \right) - \log \Gamma\left( \frac{\nu + m}{2} \right) \\
    &+ \frac{\nu + m}{2n_k} \sum_{i=1}^{n_k} \log\left( 1 + \frac{(\mb{x}_i-\hat{\bs{\mu}}_k)^T \hat{\bs{\Sigma}}_k^{-1}(\mb{x}_i-\hat{\bs{\mu}}_k)}{\nu} \right) \\
    k =& \arg \min_{k \in [1,K]} \left(1+\frac{\hat{\nu}_k}{m}\right) \log \left(1 + \frac{(\mb{x}_i-\hat{\bs{\mu}}_k)^T \hat{\bs{\Sigma}}_k^{-1}(\mb{x}_i-\hat{\bs{\mu}}_k)}{\hat{\nu}_k} \right) \\
    &+ \log(\hat{\nu}_k) + \frac{1}{m} \log\left(|\hat{\bs{\Sigma}}_k|\right) + \frac{2}{m} \log \left( \frac{ \Gamma\left(\frac{\hat{\nu}_k}{2}\right)}{\Gamma\left(\frac{\hat{\nu}_k+m}{2}\right)} \right).
\end{align*}

\textit{Proof}: See Section 9

\subsection{Implementation details and numerical considerations}

The structure of the proposed algorithm is similar to the t-QDA or RQDA algorithm. It involves recursively estimating the parameters of each cluster and then classifying test observations based on the derived decision rule.

\begin{algorithm}[!t]
\caption{FEMDA algorithm} \label{alg:CV1}
\textbf{Inputs}: 
\begin{itemize}
    \item $\{\mb{x}_i\}_{{i=1}^n_{tr}} \in \mathbb{R}^m$ observations in train set
    \item $\{y_i\}_{{i=1}^n_{tr}} \in \{1, ..., K\}$ the train labels
    \item $\{\mb{x}_i\}_{{i=1}^n_{te}} \in \mathbb{R}^m$ observations in test set
\end{itemize}
\textbf{Output}: $\{y_i\}_{{i=1}^n_{te}}$ the test labels \\
\textbf{Hyperparameters}:
\begin{itemize}
    \item $N_{iter}$ maximum number of iterations for the estimation
    \item $\epsilon$ related to the condition of convergence
    \item $\lambda$ regularization parameter
\end{itemize}
\begin{algorithmic}[1]
\For{$k \in [1, K]$}
    \State $it, convergence \longleftarrow 0, False$   
    \State $\bs{\mu}_k^0, \bs{\Sigma}_k^0 = \bs{\mu}_k^{MLE}, \bs{\Sigma}_k^{MLE} + \lambda I_m$, initialization with the regularized MLE estimator
    \While{not $convergence$ \textbf{and} $it < N_{iter}$}
        \For{$i \in [1, n_k]$}
            \State Compute $w_{i,k} = min\left(0.5, \frac{1}{(\mb{x}_i-\bs{\mu}_k^{it})^T\bs{\Sigma}_k^{-1 it}(\mb{x}_i-\bs{\mu}_k^{it})}\right)$
        \EndFor
        \State $\bs{\mu}_k^{it+1} = \frac{\sum_{i=1}^{n_k} w_{i,k} \mb{x}_i}{\sum_{i=1}^{n_k} w_{i,k}}$
        \State $\bs{\Sigma}_k^{it+1} = \frac{m}{n_k} \sum_{i=1}^{n_k} w_{i,k} (\mb{x}_i-\bs{\mu}_k^{it})(\mb{x}_i-\bs{\mu}_k^{it})^T$
        \State Regularization: $\bs{\Sigma}_k^{it+1} = \bs{\Sigma}_k^{it+1} + \lambda I_m$
        \State $it, convergence \longleftarrow it + 1, ||\bs{\Sigma}_k^{it+1} - \bs{\Sigma}_k^{it}||_1 + ||\bs{\mu}_k^{it+1} - \bs{\mu}_k^{it}||_1 < \epsilon$ 
    \EndWhile
\EndFor
\For{$i \in [1, n_{te}]$}
    \For{$k \in [1, K]$}
        \State Compute the score $s_{i,k} = \log\left((\mb{x}_i-\hat{\bs{\mu}}_k)^T\hat{\bs{\Sigma}}_k^{-1}(\mb{x}_i-\hat{\bs{\mu}}_k)\right) + \frac{1}{m} \log\left(|\hat{\bs{\Sigma}}_k|\right)$
    \EndFor
    \State Label the observation $\mb{x}_i$ with $y_i = \arg \min_{k \in [1, K]} s_{i,k}$
\EndFor
\end{algorithmic}
\end{algorithm}

In our algorithm, the estimated parameters are initialized using a regularized MLE. This helps to obtain more stable and reliable initial parameter values. Additionally, to avoid singularity problems, we regularize the estimated scatter matrix at each iteration. Inspired by Huber \cite{Huber1977Robust}, we apply a trimming to avoid diverging weights that occur when the estimated mean gets too close to a training observation. The last point that needs to be discussed is the convergence of this fixed-point equation. In 1976, Maronna \cite{Maronna1976Robust} studied the convergence of fixed-point equations defining M-estimators and proved it under restrictive assumptions that are not fulfilled here. We sometimes encounter convergence issues during our experiments, which suggests that convergence is not guaranteed here. To address this problem, we limit the number of iterations when solving the equation. This algorithm ends up being close to the robust version of classic QDA, except that we compare the logarithm of the squared Mahalanobis distances rather than the squared Mahalanobis distances. 

\section{Experiments on synthetical data}

\subsection{Classification methods used for comparison}

To evaluate the performances of the FEMDA algorithm, we perform classification tasks on various simulated and real datasets. We compare the performances with other discriminant analysis methods and more machine learning-based techniques. All the results presented are averaged over $N$ simulations. We shuffle the data in each simulation and randomly select a new train and test set. We allocate a comparable training time for each method to ensure a fair comparison. Since some methods are intrinsically faster than others, we gradually reduce the size of the train set until the training time reaches at most $T$ times the fastest method's training time. The goal of such a process is to compare the performances of the different methods for real-time applications where the trade-off between computation time and performance is crucial. We choose $N=5$ and $T=30$.

\subsubsection{Discriminant analysis methods}

During the experiments, FEMDA will be compared to the methods recalled above and the robust version of QDA implemented with Tyler's M-estimators \cite{Maronna1976Robust}. To avoid convergence issues, singular matrix estimations, and diverging weights, we apply the respective following modifications: 
\begin{itemize}
    \item For fixed-point equations solving, we limit the number of iterations to $N_{iter}=10$ and choose for the convergence condition $\epsilon=10^{-5}$,
    \item We regularize the estimated matrix with $\lambda=10^{-5}$,
    \item We use the trimmed version $\Tilde{w}_{ik} = min \left( 0.5, w_{ik} \right)$.
\end{itemize}

\subsubsection{Machine learning methods}

We also compare our performances with Support Vector Classifier (SVC) \cite{Brereton2010Support}, $k$-Nearest Neighbors (KNN) \cite{Cover1967Nearest}, and Random Forest (RF) \cite{Dess2013Enhancing} algorithms. The train set used by discriminant analysis methods for parameter estimation is split into a smaller train set and a validation set for the machine learning methods. We then perform a hyperparameter optimization using the Sklearn \cite{scikit-learn} RandomizedGridSearchCV function on the validation set. We use the following sets of values to perform the optimization: 

\vspace{0.4cm}

\resizebox{11.3cm}{!}{
\begin{tabular}{|p{2.2cm}|p{2cm}|p{2cm}|p{2cm}|p{2cm}|p{2cm}|}
 \hline
\multicolumn{6}{|c|}{Hyperparameter optimization} \\
 \hline
 \hline
\multicolumn{2}{|c|}{\textbf{SVC}} & \multicolumn{2}{c|}{\textbf{KNN}} & \multicolumn{2}{c|}{\textbf{RF}}\\
 \hline
 regularization & $[0.01, 0.1, 1]$ & Nb neighbors & $[5, 11, 15, 21]$ & Nb trees & $[50, 100, 200]$ \\
 \hline
 kernel function & [rbf, poly, sigm] & Tree leaf size & $[20, 30, 40]$ & Max depth & $[5, 10, 15, 20]$ \\
  \hline
  kernel parameter & $[0.01, 0.1, 1]$ & Minkowski power & $[1, 2, 3]$ & Min sample split & $[2, 4, 5]$ \\
 \hline
  \cellcolor{thisisgrey} & \cellcolor{thisisgrey} & \cellcolor{thisisgrey} & \cellcolor{thisisgrey} &  Max features & $[1, 2, 3]$ \\
\hline
 \cellcolor{thisisgrey} & \cellcolor{thisisgrey} & \cellcolor{thisisgrey} & \cellcolor{thisisgrey} & Max leaf nodes & $[None, 4, 5]$ \\
\hline
Nb samples & $10$ & Nb samples & $20$ & Nb samples & $10$ \\
 \hline
\end{tabular}
}

\subsection{Data generation}

Our main goal is to compare the performances of different discriminant analysis methods. Since all these methods assume that clusters have an underlying specific statistical model, we will support such a hypothesis by generating the data with one elliptically symmetric distribution per cluster. Thus, for a given number of points $n$ splitted in clusters $K$ with priors $\pi_1,...,\pi_K$, we need to construct randomly $K$ covariance matrix $\bs{\Sigma}_1,...,\bs{\Sigma}_K \in \mathbb{R}^{m \times m}$, mean vectors $\bs{\mu}_1,...,\bs{\mu}_K \in \mathbb{R}^m$ and define the density generator functions $g_1,...,g_K$ that will then be used to generate the points.

\subsubsection{Mean vector random generation}

For each $\bs{\mu}_i$, we generate a random point on the $m$-sphere of radius $r$. Regarding the norm of the covariance matrix, we choose a radius that, on the one hand, does not make the classification too easy (clusters too far away from one another) and, on the other hand, ensures that it remains possible (clusters do not overlap too much).

\subsubsection{Covariance matrix random generation}

The generation of the covariance matrix is trickier as poor choices for the eigenvalues will significantly impact the methods' performances. While a too-large eigenvalue is completely non-informative and is useless for the classification, a too-small eigenvalue will be too discriminant. To generate $\bs{\Sigma}_i$, we first generate $\lambda_{i1},...,\lambda_{im}$, then a random rotation matrix $P_i$ and finally compute $\bs{\Sigma}_i = P_i Diag(\lambda_{i1},...,\lambda_{im})P_i^{-1}$.

The random rotation matrix $P_i$ is drawn using the Haar distribution \cite{stewart1980efficient} implemented in Scipy.

We generate the eigenvalues using a $\chi^2$ distribution trimmed by a minimum and maximum value. The higher the degree of freedom of the $\chi_2$ distribution is, the less informative the eigenvalues become. To get some heterogeneity in the clusters, we choose a random degree of freedom using a Poisson distribution of parameter $\xi$:
\begin{align*}
    \lambda_{ij} \sim min\left( max\left( \chi^2\left(P(\xi)\right), \lambda_{min} \right), \lambda_{max} \right).
\end{align*}

In practice, we choose:
\begin{itemize}
    \item $\xi=1$,
    \item $\lambda_{min}=1$,
    \item $\lambda_{max}=20$.
\end{itemize}

\subsubsection{Density generator function definition}

To generate the data of each cluster, we choose the density generator function among two parametric families:
\begin{itemize}
    \item Multivariate generalized Gaussian distributions \cite{Pascal2013Parameter} of shape parameter $\beta$,
    \item Multivariate student distributions \cite{Roth2012On} with degree of freedom $\nu$.
\end{itemize}

\subsection{Performances of the estimators}

We start by comparing the training time of each method. We display both the training time and the percentage of data effectively exploited during the training phase. Since the percentage of data used is gradually decreased to ensure a similar training time for all methods, a lower percentage of processed data necessarily implies a slower method.

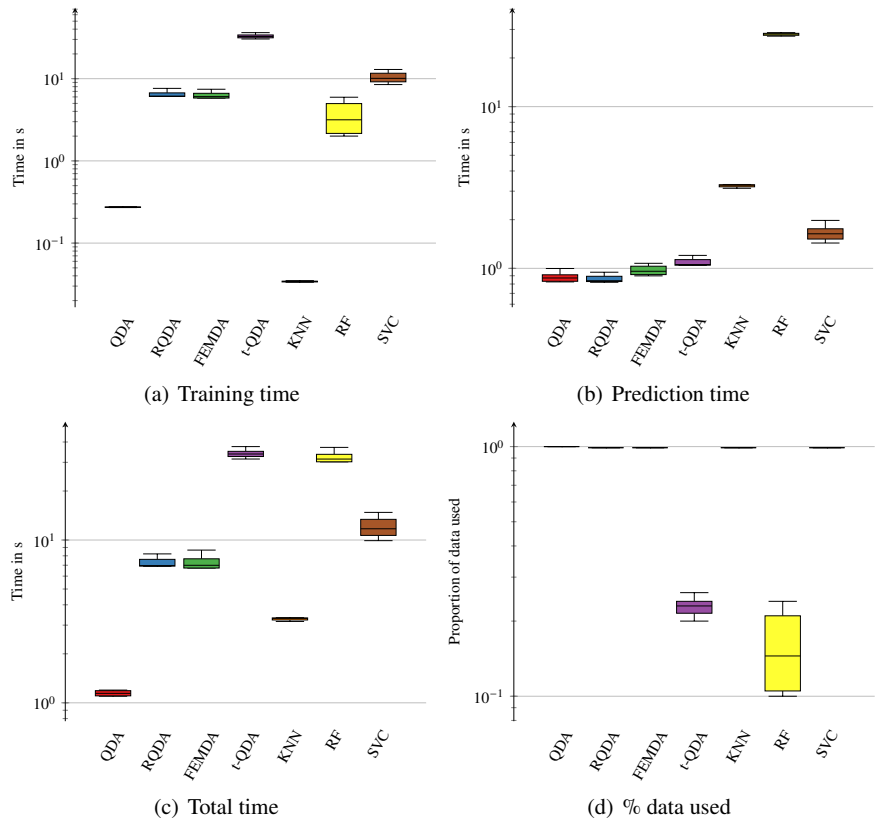
\begin{figure}[H]
\centering
\subfigure[Training time]{
\begin{tikzpicture}[scale=0.70]
	\pgfplotstableread[col sep=comma]{data_simulated_training_time.csv}\csvdata
	\pgfplotstabletranspose\datatransposed{\csvdata} 
	\begin{axis}[
		boxplot/draw direction = y,
		x axis line style = {opacity=0},
		axis x line* = bottom,
		axis y line = left,
		enlarge y limits,
		ymajorgrids,
		xtick = {1, 2, 3, 4, 5, 6, 7},
		xticklabel style = {align=center, font=\small, rotate=60},
		xticklabels = {QDA, RQDA, FEMDA, t-QDA, KNN, RF, SVC},
		xtick style = {draw=none},
		ylabel = {Time in s},
        ymode=log
	]
		\foreach \n in {1,...,7} {
			\addplot+[boxplot, fill, draw=black] table[y index=\n] {\datatransposed};
		}
	\end{axis}
\end{tikzpicture}
}
\subfigure[Prediction time]{
\begin{tikzpicture}[scale=0.70]
	\pgfplotstableread[col sep=comma]{data_simulated_testing_time.csv}\csvdata
	\pgfplotstabletranspose\datatransposed{\csvdata} 
	\begin{axis}[
		boxplot/draw direction = y,
		x axis line style = {opacity=0},
		axis x line* = bottom,
		axis y line = left,
		enlarge y limits,
		ymajorgrids,
		xtick = {1, 2, 3, 4, 5, 6, 7},
		xticklabel style = {align=center, font=\small, rotate=60},
		xticklabels = {QDA, RQDA, FEMDA, t-QDA, KNN, RF, SVC},
		xtick style = {draw=none},
		ylabel = {Time in s},
        ymode=log
	]
		\foreach \n in {1,...,7} {
			\addplot+[boxplot, fill, draw=black] table[y index=\n] {\datatransposed};
		}
	\end{axis}
\end{tikzpicture}
}
\subfigure[Total time]{
\begin{tikzpicture}[scale=0.70]
	\pgfplotstableread[col sep=comma]{data_simulated_total_time.csv}\csvdata
	\pgfplotstabletranspose\datatransposed{\csvdata} 
	\begin{axis}[
		boxplot/draw direction = y,
		x axis line style = {opacity=0},
		axis x line* = bottom,
		axis y line = left,
		enlarge y limits,
		ymajorgrids,
		xtick = {1, 2, 3, 4, 5, 6, 7},
		xticklabel style = {align=center, font=\small, rotate=60},
		xticklabels = {QDA, RQDA, FEMDA, t-QDA, KNN, RF, SVC},
		xtick style = {draw=none},
		ylabel = {Time in s},
        ymode=log
	]
		\foreach \n in {1,...,7} {
			\addplot+[boxplot, fill, draw=black] table[y index=\n] {\datatransposed};
		}
	\end{axis}
\end{tikzpicture}
}
\subfigure[\% data used]{
\begin{tikzpicture}[scale=0.70]
	\pgfplotstableread[col sep=comma]{data_simulated_data_used.csv}\csvdata
	\pgfplotstabletranspose\datatransposed{\csvdata} 
	\begin{axis}[
		boxplot/draw direction = y,
		x axis line style = {opacity=0},
		axis x line* = bottom,
		axis y line = left,
		enlarge y limits,
		ymajorgrids,
		xtick = {1, 2, 3, 4, 5, 6, 7},
		xticklabel style = {align=center, font=\small, rotate=60},
		xticklabels = {QDA, RQDA, FEMDA, t-QDA, KNN, RF, SVC},
		xtick style = {draw=none},
		ylabel = {Proportion of data used},
        ymode=log
	]
		\foreach \n in {1,...,7} {
			\addplot+[boxplot, fill, draw=black] table[y index=\n] {\datatransposed};
		}
	\end{axis}
\end{tikzpicture}
}
\caption{Training and testing computation times comparison between the different algorithms, and study of the percentage of data used for training on simulated dataset}
\label{fig:1} 
\end{figure}

Among the discriminant analysis methods, QDA is the fastest. Indeed, FEMDA, RQDA, and t-QDA all involve solving iteratively a fixed-point equation for the estimation part, which takes more time. FEMDA and RQDA have similar computation times since their fixed-point equation have close expressions. However, t-QDA is significantly slower because it requires solving an optimization problem at each iteration of the fixed-point equation. Despite using only a fraction of the available training data, t-QDA is the slowest method to train. The prediction time is roughly the same for all discriminant analysis methods, as it always consists in evaluating the likelihood using the estimated parameters. For machine learning methods, it is relevant to display the prediction time as some methods may be fast to train but slow to predict (such as KNN). When comparing the total time, which includes both training and prediction time, machine learning methods, on average, tend to be slower than discriminant analysis methods. Among all the methods, only QDA and KNN are faster than FEMDA. In particular, FEMDA is ten times faster than t-QDA while processing three times more data.

Since we know the mean and covariance matrix of the elliptically symmetric distributions used to generate the data, we can compare the accuracy of the different estimators. For the mean vector, we compute the Relative Mean Square Error (RMSE) between the real and the estimated parameter: $error(\bs{\mu}) = \frac{|| \bs{\mu} - \bs{\hat{\mu}}||}{|| \bs{\mu} ||}$. This error is then averaged across the different clusters. However, such a metric is less relevant for the covariance matrix as FEMDA's fixed point equation does not even impose a specific scale on the solution. A more suitable metric for comparing the covariance matrix estimates is the Riemannian distance over the positive-definite matrix \cite{Shahbazi2021Using}. Once again, we average the Riemannian distance across the different clusters.

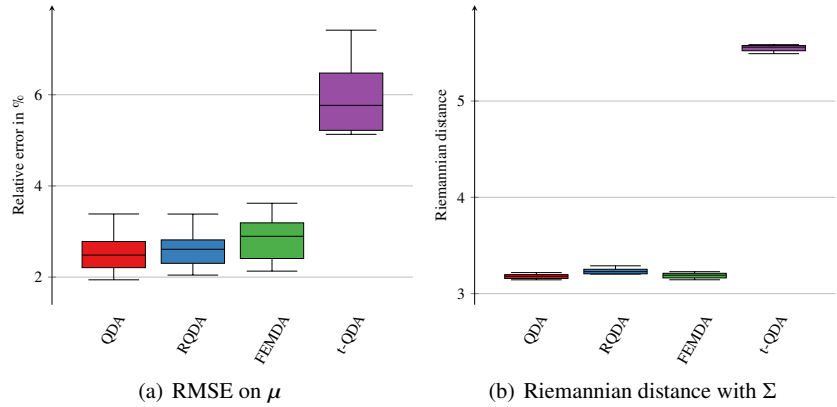
\begin{figure}[H]
\centering
\subfigure[RMSE on $\mu$]{
\begin{tikzpicture}[scale=0.70]
	\pgfplotstableread[col sep=comma]{data_simulated_mu_error.csv}\csvdata
	\pgfplotstabletranspose\datatransposed{\csvdata} 
	\begin{axis}[
		boxplot/draw direction = y,
		x axis line style = {opacity=0},
		axis x line* = bottom,
		axis y line = left,
		enlarge y limits,
		ymajorgrids,
		xtick = {1, 2, 3, 4},
		xticklabel style = {align=center, font=\small, rotate=60},
		xticklabels = {QDA, RQDA, FEMDA, t-QDA},
		xtick style = {draw=none},
		ylabel = {Relative error in \%},
	]
		\foreach \n in {1,...,4} {
			\addplot+[boxplot, fill, draw=black] table[y index=\n] {\datatransposed};
		}
	\end{axis}
\end{tikzpicture}
}
\subfigure[Riemannian distance with $\Sigma$]{
\begin{tikzpicture}[scale=0.70]
	\pgfplotstableread[col sep=comma]{data_simulated_Sigma_error.csv}\csvdata
	\pgfplotstabletranspose\datatransposed{\csvdata} 
	\begin{axis}[
		boxplot/draw direction = y,
		x axis line style = {opacity=0},
		axis x line* = bottom,
		axis y line = left,
		enlarge y limits,
		ymajorgrids,
		xtick = {1, 2, 3, 4},
		xticklabel style = {align=center, font=\small, rotate=60},
		xticklabels = {QDA, RQDA, FEMDA, t-QDA},
		xtick style = {draw=none},
		ylabel = {Riemannian distance},
	]
		\foreach \n in {1,...,4} {
			\addplot+[boxplot, fill, draw=black] table[y index=\n] {\datatransposed};
		}
	\end{axis}
\end{tikzpicture}
}
\caption{Estimation error comparison between the estimators of the different discriminant analysis methods}
\label{fig:2} 
\end{figure}

The relative error on the estimation of the mean vector is approximately the same for QDA, RQDA, and FEMDA, at around 3\%. t-QDA, however, happens to be the least accurate, with a relative error exceeding 6\%, which is twice as high as the other methods. Similarly, when considering the estimation of the covariance matrix, QDA, RQDA, and FEMDA obtain very similar average Riemannian distances to $\Sigma$, around $3.2$. Again, t-QDA's estimator is the least accurate, with an average distance of around $5.6$. Given that t-QDA already had a lower accuracy on the mean vector, its lower performance in estimating the covariance matrix is expected.

\subsection{Performances on clean data}

We now compare the classification performances of the different methods on test data simulated in the same way. Data generation parameters are the following: 

\vspace{0.4cm}

\resizebox{11.3cm}{!}{
\begin{tabular}{|p{4cm}|p{8cm}|}
 \hline
Number of clusters & 3 \\
\hline
Dimension & 10 \\
\hline
Number of points & 3000 \\
\hline
Radius for $\mu$ generation & 2 \\
\hline
$\lambda$ parameter for eigenvalues & 1 \\
\hline
Minimum eigenvalue value & 1 \\
\hline
Maximum eigenvalue value & 20 \\
\hline
Cluster 1 & Prior of 0.33 and Generalized gaussian with $\beta=0.8$ \\
\hline
Cluster 2 & Prior of 0.33 and Generalized gaussian with $\beta=1.5$ \\
\hline
Cluster 3 & Prior of 0.34 and multivariate student with $\nu=10$ \\
\hline
\end{tabular}
  }

\vspace{0.4cm}

We obtain the following results:

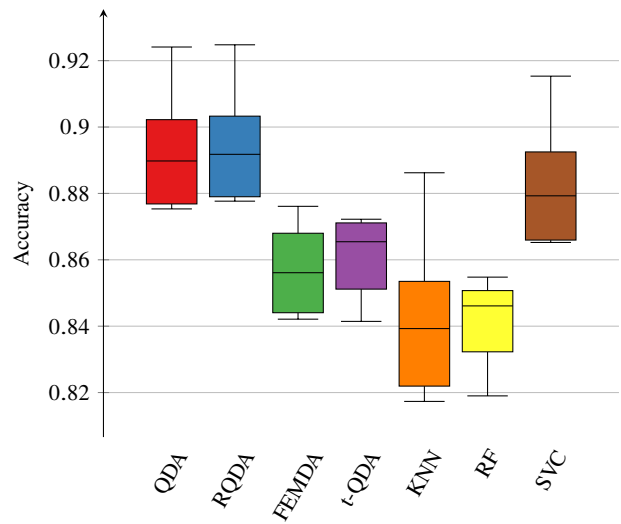
\begin{figure}[H]
\centering
\begin{tikzpicture}[scale=1]
	\pgfplotstableread[col sep=comma]{data_simulated_accuracy.csv}\csvdata
	\pgfplotstabletranspose\datatransposed{\csvdata} 
	\begin{axis}[
		boxplot/draw direction = y,
		x axis line style = {opacity=0},
		axis x line* = bottom,
		axis y line = left,
		enlarge y limits,
		ymajorgrids,
		xtick = {1, 2, 3, 4, 5, 6, 7},
		xticklabel style = {align=center, font=\small, rotate=60},
		xticklabels = {QDA, RQDA, FEMDA, t-QDA, KNN, RF, SVC},
		xtick style = {draw=none},
		ylabel = {Accuracy},
	]
		\foreach \n in {1,...,7} {
			\addplot+[boxplot, fill, draw=black] table[y index=\n] {\datatransposed};
		}
	\end{axis}
\end{tikzpicture}
\caption{Classification accuracy comparison between the different machine learning methods on simulated dataset}
\label{fig:3} 
\end{figure}

Overall, discriminant analysis methods outperform machine learning methods, probably due to the elliptical structure of the simulated data. 
As there is no noise and the distributions used for each cluster are close to Gaussian distributions, QDA and RQDA are expected to perform above other methods. FEMDA performs better than KNN and RF but is below SVC and t-QDA, although the latter two methods require a longer total time for training and prediction.

\subsection{Performances on noisy data}

To conclude the study on simulated data, we will now evaluate the robustness of each method to outliers. We replace a fraction of the train set by a uniform random noise. These points are randomly generated within the hypercube containing the original train set. We focus on the evolution in accuracy as the percentage of contaminated data increases. 

\begin{figure}[H]
\centering
\subfigure{\includegraphics[width=13cm]{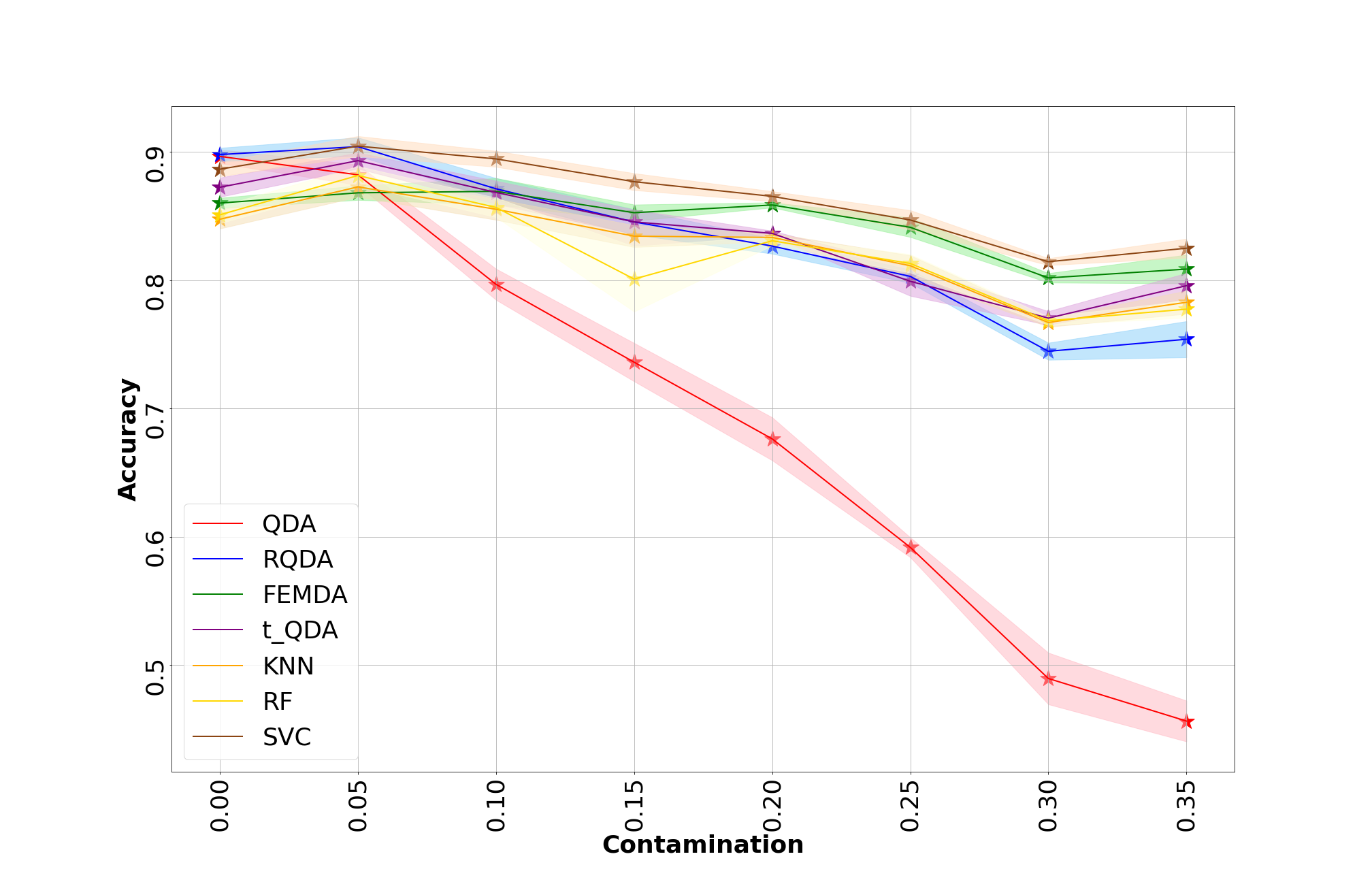}}
\caption{Accuracy evolution of the different algorithms when a percentage of the simulated train set is replaced by mild Gaussian noise}
\label{fig:4} 
\end{figure}

Intrinsically non-robust, QDA is the most affected method by the noise. Its performance decreases slightly above a random classifier at the maximum contamination rate. However, since the noise introduced is random and has no specific pattern, most methods still have sufficient data to capture each cluster's inherent structure accurately. Machine learning methods handle the contamination well, as they lose an average of 8\% accuracy at the maximum contamination rate of 35\%. Among the robust methods, RQDA is the most impacted, with a 15\% decrease in accuracy. On the other hand, FEMDA and t-QDA show more robustness and only lose around 5\% in accuracy. FEMDA is the method with the smallest performance decrease and becomes the most accurate discriminant analysis method at a 10\% 

\section{Experiments on real data}

In this section, our objective is to compare the performances of our classification methods on real datasets. Along with simulated data, we first study the accuracy of clean data before adding noise to evaluate the robustness of the methods. To optimize the performances of all methods, we perform a Principal Component Analysis (PCA) before training to reduce the input dimension. Our experiments focus on four real datasets obtained from the UCI machine learning repository \cite{UCI2017}: 

\vspace{0.4cm}

\resizebox{11.3cm}{!}{
\begin{tabular}{|p{4cm}|p{2cm}|p{2cm}|p{2cm}|p{2cm}|}
 \hline
 Name & Avila & Frogs species & Glass & Sonar \\
 \hline
 Number of data & 20867 & 7195 & 214 & 208 \\
 \hline
 \% used in train set & 70\% & 70\% & 70\% & 70\% \\
 \hline
 Dimension & 10 & 22 & 9 & 60 \\
 \hline
 Dimension after PCA & 8 & 14 & 5 & 16 \\
 \hline
 Number of clusters & 12 & 10 & 6 & 2 \\
 \hline
 Prior & [0.41, 0.00, 0.00, 0.03, 0.1, 0.18, 0.04, 0.04, 0.08, 0.04, 0.05, 0.03] & [0.09, 0.48, 0.08, 0.04, 0.07, 0.15, 0.04, 0.02, 0.01, 0.02] & [0.32, 0.35, 0.08, 0.06, 0.05, 0.14] & [0.54, 0.46] \\
 \hline
\end{tabular}
}

\vspace{0.4cm}

\begin{figure}[H]
\centering
\subfigure[Avila]{
\begin{tikzpicture}[scale=0.69]
	\pgfplotstableread[col sep=comma]{data_avila_total_time.csv}\csvdata
	\pgfplotstabletranspose\datatransposed{\csvdata} 
	\begin{axis}[
		boxplot/draw direction = y,
		x axis line style = {opacity=0},
		axis x line* = bottom,
		axis y line = left,
		enlarge y limits,
		ymajorgrids,
		xtick = {1, 2, 3, 4, 5, 6, 7},
		xticklabel style = {align=center, font=\small, rotate=60},
		xticklabels = {QDA, RQDA, FEMDA, t-QDA, KNN, RF, SVC},
		xtick style = {draw=none},
		ylabel = {Time in s},
        ymode=log
	]
		\foreach \n in {1,...,7} {
			\addplot+[boxplot, fill, draw=black] table[y index=\n] {\datatransposed};
		}
	\end{axis}
\end{tikzpicture}
}
\subfigure[Frogs species]{
\begin{tikzpicture}[scale=0.69]
	\pgfplotstableread[col sep=comma]{data_frogs_species_total_time.csv}\csvdata
	\pgfplotstabletranspose\datatransposed{\csvdata} 
	\begin{axis}[
		boxplot/draw direction = y,
		x axis line style = {opacity=0},
		axis x line* = bottom,
		axis y line = left,
		enlarge y limits,
		ymajorgrids,
		xtick = {1, 2, 3, 4, 5, 6, 7},
		xticklabel style = {align=center, font=\small, rotate=60},
		xticklabels = {QDA, RQDA, FEMDA, t-QDA, KNN, RF, SVC},
		xtick style = {draw=none},
		ylabel = {Time in s},
        ymode=log
	]
		\foreach \n in {1,...,7} {
			\addplot+[boxplot, fill, draw=black] table[y index=\n] {\datatransposed};
		}
	\end{axis}
\end{tikzpicture}
}
\subfigure[Glass]{
\begin{tikzpicture}[scale=0.69]
	\pgfplotstableread[col sep=comma]{data_glass_total_time.csv}\csvdata
	\pgfplotstabletranspose\datatransposed{\csvdata} 
	\begin{axis}[
		boxplot/draw direction = y,
		x axis line style = {opacity=0},
		axis x line* = bottom,
		axis y line = left,
		enlarge y limits,
		ymajorgrids,
		xtick = {1, 2, 3, 4, 5, 6, 7},
		xticklabel style = {align=center, font=\small, rotate=60},
		xticklabels = {QDA, RQDA, FEMDA, t-QDA, KNN, RF, SVC},
		xtick style = {draw=none},
		ylabel = {Time in s},
        ymode=log
	]
		\foreach \n in {1,...,7} {
			\addplot+[boxplot, fill, draw=black] table[y index=\n] {\datatransposed};
		}
	\end{axis}
\end{tikzpicture}
}
\subfigure[Sonar]{
\begin{tikzpicture}[scale=0.69]
	\pgfplotstableread[col sep=comma]{data_sonar_total_time.csv}\csvdata
	\pgfplotstabletranspose\datatransposed{\csvdata} 
	\begin{axis}[
		boxplot/draw direction = y,
		x axis line style = {opacity=0},
		axis x line* = bottom,
		axis y line = left,
		enlarge y limits,
		ymajorgrids,
		xtick = {1, 2, 3, 4, 5, 6, 7},
		xticklabel style = {align=center, font=\small, rotate=60},
		xticklabels = {QDA, RQDA, FEMDA, t-QDA, KNN, RF, SVC},
		xtick style = {draw=none},
		ylabel = {Time in s},
        ymode=log
	]
		\foreach \n in {1,...,7} {
			\addplot+[boxplot, fill, draw=black] table[y index=\n] {\datatransposed};
		}
	\end{axis}
\end{tikzpicture}
}
\caption{Comparison of the time required by each method for training and testing on 4 real datasets}
\label{fig:5} 
\end{figure}
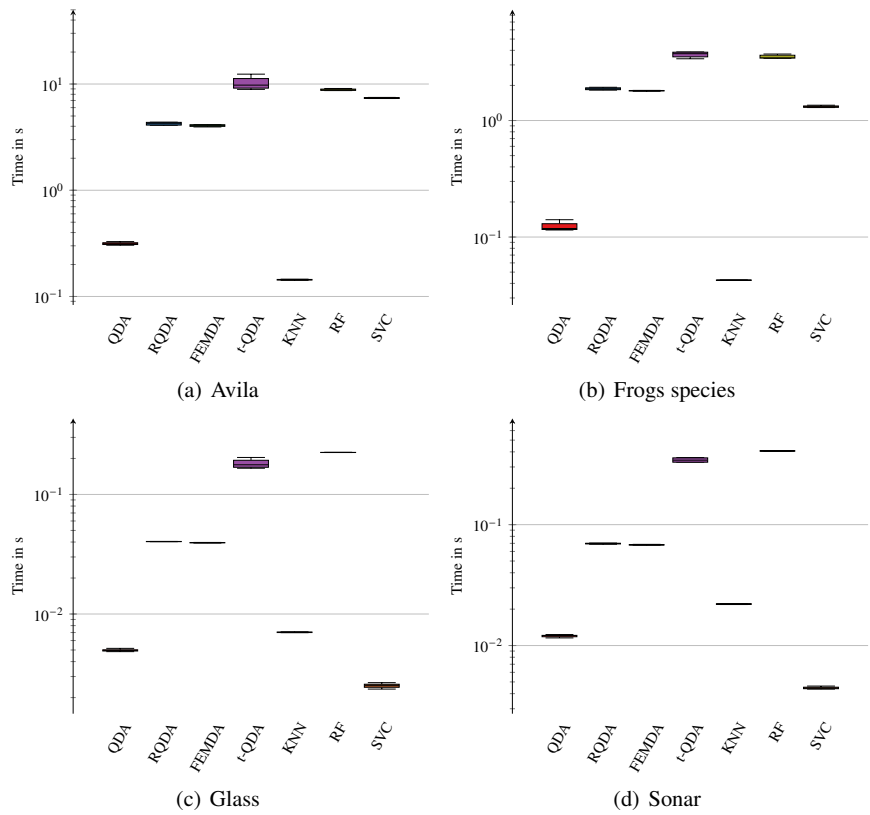

\begin{figure}[H]
\centering
\subfigure[Avila]{
\begin{tikzpicture}[scale=0.69]
	\pgfplotstableread[col sep=comma]{data_avila_data_used.csv}\csvdata
	\pgfplotstabletranspose\datatransposed{\csvdata} 
	\begin{axis}[
		boxplot/draw direction = y,
		x axis line style = {opacity=0},
		axis x line* = bottom,
		axis y line = left,
		enlarge y limits,
		ymajorgrids,
		xtick = {1, 2, 3, 4, 5, 6, 7},
		xticklabel style = {align=center, font=\small, rotate=60},
		xticklabels = {QDA, RQDA, FEMDA, t-QDA, KNN, RF, SVC},
		xtick style = {draw=none},
		ylabel = {Proportion of data used},
        ymode=log
	]
		\foreach \n in {1,...,7} {
			\addplot+[boxplot, fill, draw=black] table[y index=\n] {\datatransposed};
		}
	\end{axis}
\end{tikzpicture}
}
\subfigure[Frogs species]{
\begin{tikzpicture}[scale=0.69]
	\pgfplotstableread[col sep=comma]{data_frogs_species_data_used.csv}\csvdata
	\pgfplotstabletranspose\datatransposed{\csvdata} 
	\begin{axis}[
		boxplot/draw direction = y,
		x axis line style = {opacity=0},
		axis x line* = bottom,
		axis y line = left,
		enlarge y limits,
		ymajorgrids,
		xtick = {1, 2, 3, 4, 5, 6, 7},
		xticklabel style = {align=center, font=\small, rotate=60},
		xticklabels = {QDA, RQDA, FEMDA, t-QDA, KNN, RF, SVC},
		xtick style = {draw=none},
		ylabel = {Proportion of data used},
        ymode=log
	]
		\foreach \n in {1,...,7} {
			\addplot+[boxplot, fill, draw=black] table[y index=\n] {\datatransposed};
		}
	\end{axis}
\end{tikzpicture}
}
\subfigure[Glass]{
\begin{tikzpicture}[scale=0.69]
	\pgfplotstableread[col sep=comma]{data_glass_data_used.csv}\csvdata
	\pgfplotstabletranspose\datatransposed{\csvdata} 
	\begin{axis}[
		boxplot/draw direction = y,
		x axis line style = {opacity=0},
		axis x line* = bottom,
		axis y line = left,
		enlarge y limits,
		ymajorgrids,
		xtick = {1, 2, 3, 4, 5, 6, 7},
		xticklabel style = {align=center, font=\small, rotate=60},
		xticklabels = {QDA, RQDA, FEMDA, t-QDA, KNN, RF, SVC},
		xtick style = {draw=none},
		ylabel = {Proportion of data used},
        ymode=log
	]
		\foreach \n in {1,...,7} {
			\addplot+[boxplot, fill, draw=black] table[y index=\n] {\datatransposed};
		}
	\end{axis}
\end{tikzpicture}
}
\subfigure[Sonar]{
\begin{tikzpicture}[scale=0.69]
	\pgfplotstableread[col sep=comma]{data_sonar_data_used.csv}\csvdata
	\pgfplotstabletranspose\datatransposed{\csvdata} 
	\begin{axis}[
		boxplot/draw direction = y,
		x axis line style = {opacity=0},
		axis x line* = bottom,
		axis y line = left,
		enlarge y limits,
		ymajorgrids,
		xtick = {1, 2, 3, 4, 5, 6, 7},
		xticklabel style = {align=center, font=\small, rotate=60},
		xticklabels = {QDA, RQDA, FEMDA, t-QDA, KNN, RF, SVC},
		xtick style = {draw=none},
		ylabel = {Proportion of data used},
        ymode=log
	]
		\foreach \n in {1,...,7} {
			\addplot+[boxplot, fill, draw=black] table[y index=\n] {\datatransposed};
		}
	\end{axis}
\end{tikzpicture}
}
\caption{Comparison of the percentage of data used by each method for training and testing on four real datasets}
\label{fig:6} 
\end{figure}
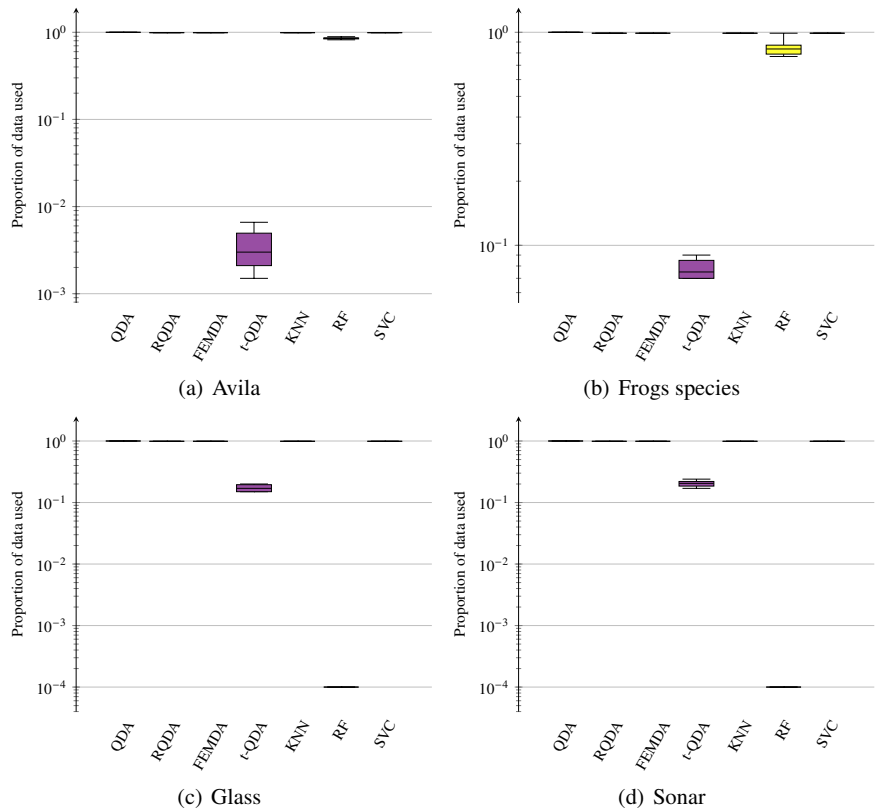

The computation time analysis is roughly the same as for simulated data. QDA is the fastest method, while t-QDA is the slowest. Again, FEMDA and RQDA exhibit similar computation times, with a slight advantage for FEMDA. When dealing with large datasets, KNN tends to be the fastest method, whereas SVC becomes slower. However, SVC proves to be the fastest method for smaller datasets. Concerning the percentage of data used, all methods, except for t-QDA and RF, use almost the entire train set. On large datasets, t-QDA leaves out most of the training data, respectively more than 90\% and 99\% being left out for the Frog species and Avila datasets. On the contrary, RF is close to exploiting all the available data. However, the situation is reversed for small datasets. t-QDA uses a larger portion of the train set while RF's prediction time becomes too high to even consider using the train set, which is thus unused by RF.

\begin{figure}[H]
\centering
\subfigure[Avila]{
\begin{tikzpicture}[scale=0.69]
	\pgfplotstableread[col sep=comma]{data_avila_accuracy.csv}\csvdata
	\pgfplotstabletranspose\datatransposed{\csvdata} 
	\begin{axis}[
		boxplot/draw direction = y,
		x axis line style = {opacity=0},
		axis x line* = bottom,
		axis y line = left,
		enlarge y limits,
		ymajorgrids,
		xtick = {1, 2, 3, 4, 5, 6, 7},
		xticklabel style = {align=center, font=\small, rotate=60},
		xticklabels = {QDA, RQDA, FEMDA, t-QDA, KNN, RF, SVC},
		xtick style = {draw=none},
		ylabel = {Accuracy},
	]
		\foreach \n in {1,...,7} {
			\addplot+[boxplot, fill, draw=black] table[y index=\n] {\datatransposed};
		}
	\end{axis}
\end{tikzpicture}
}
\subfigure[Frogs species]{
\begin{tikzpicture}[scale=0.69]
	\pgfplotstableread[col sep=comma]{data_frogs_species_accuracy.csv}\csvdata
	\pgfplotstabletranspose\datatransposed{\csvdata} 
	\begin{axis}[
		boxplot/draw direction = y,
		x axis line style = {opacity=0},
		axis x line* = bottom,
		axis y line = left,
		enlarge y limits,
		ymajorgrids,
		xtick = {1, 2, 3, 4, 5, 6, 7},
		xticklabel style = {align=center, font=\small, rotate=60},
		xticklabels = {QDA, RQDA, FEMDA, t-QDA, KNN, RF, SVC},
		xtick style = {draw=none},
		ylabel = {Accuracy},
	]
		\foreach \n in {1,...,7} {
			\addplot+[boxplot, fill, draw=black] table[y index=\n] {\datatransposed};
		}
	\end{axis}
\end{tikzpicture}
}
\subfigure[Glass]{
\begin{tikzpicture}[scale=0.69]
	\pgfplotstableread[col sep=comma]{data_glass_accuracy.csv}\csvdata
	\pgfplotstabletranspose\datatransposed{\csvdata} 
	\begin{axis}[
		boxplot/draw direction = y,
		x axis line style = {opacity=0},
		axis x line* = bottom,
		axis y line = left,
		enlarge y limits,
		ymajorgrids,
		xtick = {1, 2, 3, 4, 5, 6, 7},
		xticklabel style = {align=center, font=\small, rotate=60},
		xticklabels = {QDA, RQDA, FEMDA, t-QDA, KNN, RF, SVC},
		xtick style = {draw=none},
		ylabel = {Accuracy},
	]
		\foreach \n in {1,...,7} {
			\addplot+[boxplot, fill, draw=black] table[y index=\n] {\datatransposed};
		}
	\end{axis}
\end{tikzpicture}
}
\subfigure[Sonar]{
\begin{tikzpicture}[scale=0.69]
	\pgfplotstableread[col sep=comma]{data_sonar_accuracy.csv}\csvdata
	\pgfplotstabletranspose\datatransposed{\csvdata} 
	\begin{axis}[
		boxplot/draw direction = y,
		x axis line style = {opacity=0},
		axis x line* = bottom,
		axis y line = left,
		enlarge y limits,
		ymajorgrids,
		xtick = {1, 2, 3, 4, 5, 6, 7},
		xticklabel style = {align=center, font=\small, rotate=60},
		xticklabels = {QDA, RQDA, FEMDA, t-QDA, KNN, RF, SVC},
		xtick style = {draw=none},
		ylabel = {Accuracy},
	]
		\foreach \n in {1,...,7} {
			\addplot+[boxplot, fill, draw=black] table[y index=\n] {\datatransposed};
		}
	\end{axis}
\end{tikzpicture}
}
\caption{Classification accuracy comparison between the different machine learning methods on four real datasets}
\label{fig:7} 
\end{figure}
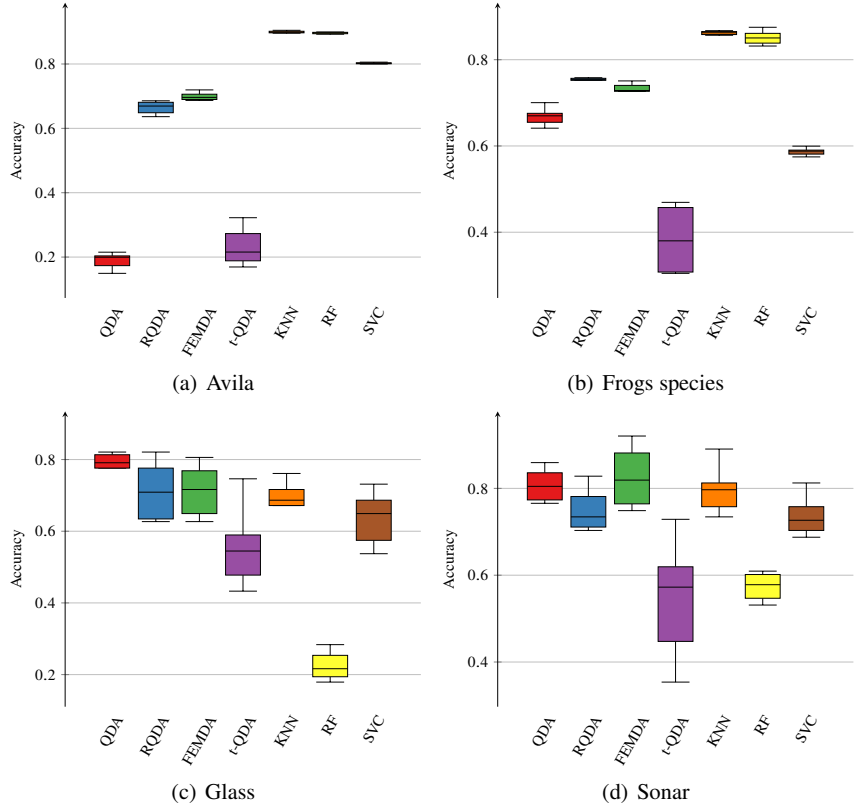

Despite training involving all data on Avila, QDA is slightly above t-QDA, indicating that this dataset's structure is far from Gaussianity. FEMDA and RQDA, however, perform well on Avila and Frogs species, their robustness allowing them to handle the non-Gaussianity better. As expected for big datasets, machine learning methods outperform discriminant analysis methods. We observe the opposite on smaller datasets: discriminant analysis methods perform better than machine learning methods, probably because the underlying statistical model compensates for the lack of data. The terrible performances of RF were expected since no training data could be used. On Glass and Sonar datasets, QDA  retrieves good performances, while t-QDA is still penalized by the low amount of processed data during the training part. Again, RQDA and FEMDA obtain very similar performances, FEMDA being the best method on the Sonar dataset.

On both Avila and Frogs species, t-QDA performs very poorly because of the very low amount of data processed during training. Despite training involving all data, QDA is only slightly above t-QDA, indicating that these datasets' structures are far from Gaussianity. In both cases, FEMDA is the best discriminant analysis method but gets overwhelmed by machine learning methods. On the smaller datasets, Glass and Sonar, t-QDA has access to more data for the training part, which leads to much better performances. It is consistent with the fact that it is considered the best discriminant analysis method when the training time is not restricted. However, it still remains the least efficient discriminant analysis method in this framework. As expected with the inexistent training part, RF has a terrible performance on Glass. The two other machine learning methods perform well but still get outperformed by QDA, RQDA, and FEMDA on low sample sizes.

Finally, to conclude the experiments, we contaminate the training data with random noise like for simulated data. 

\begin{figure}[H]
\centering
\subfigure[Avila\label{fig:8a}]{\includegraphics[width=5.8cm]{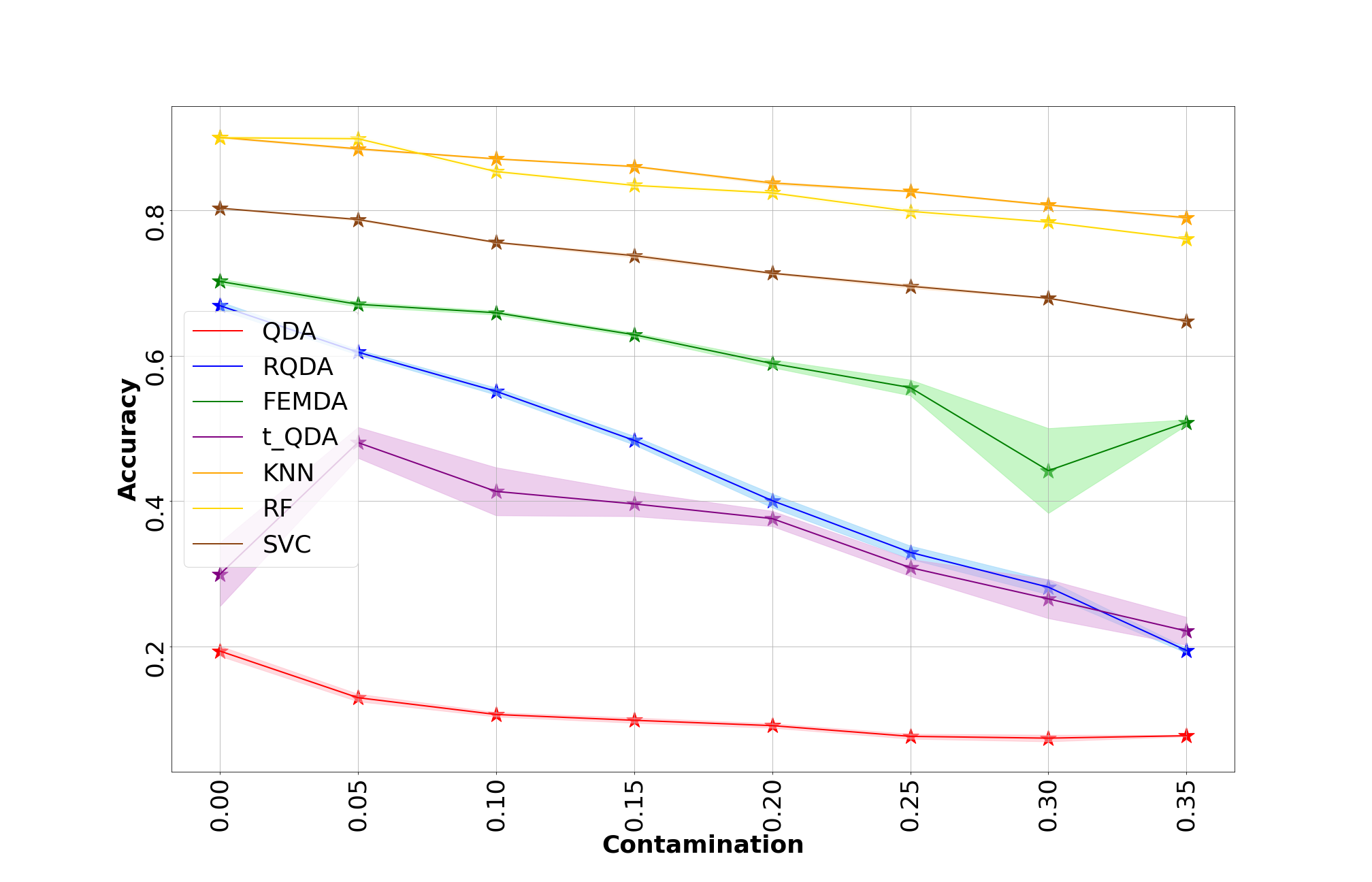}}
\subfigure[Frogs species\label{fig:8b}]{\includegraphics[width=5.8cm]{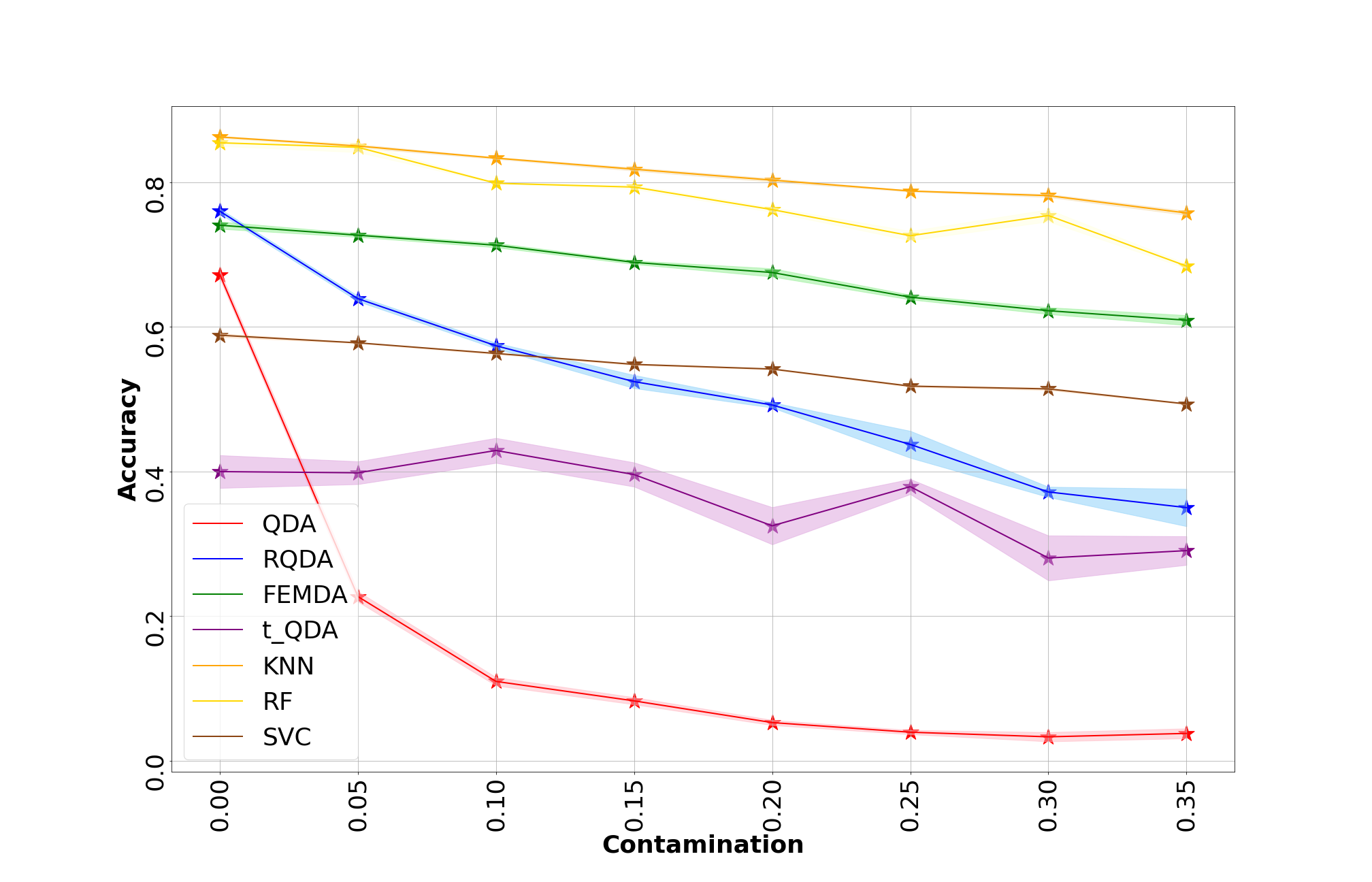}}
\subfigure[Glass\label{fig:8c}]{\includegraphics[width=5.8cm]{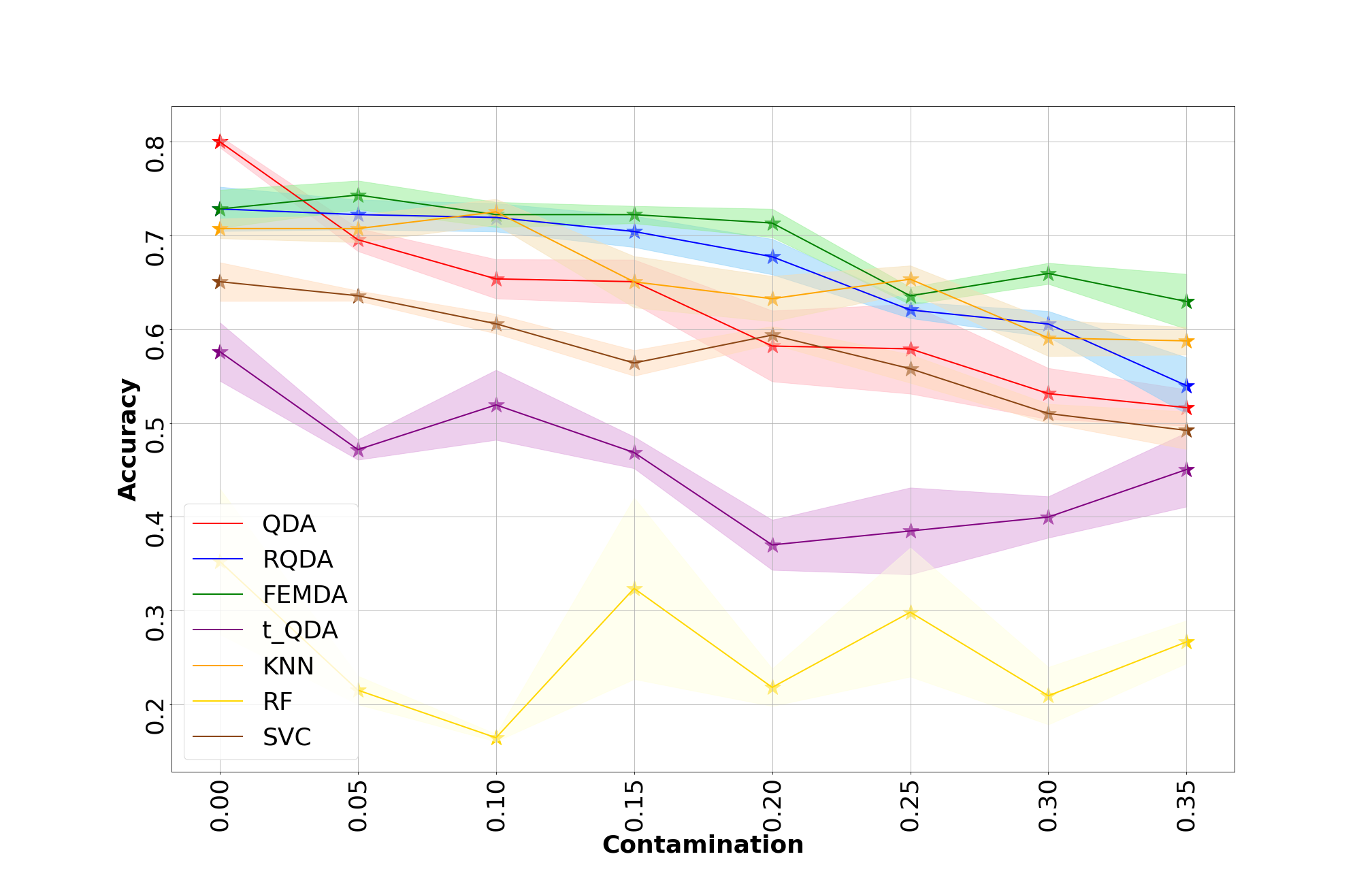}}
\subfigure[Sonar\label{fig:8d}]{\includegraphics[width=5.8cm]{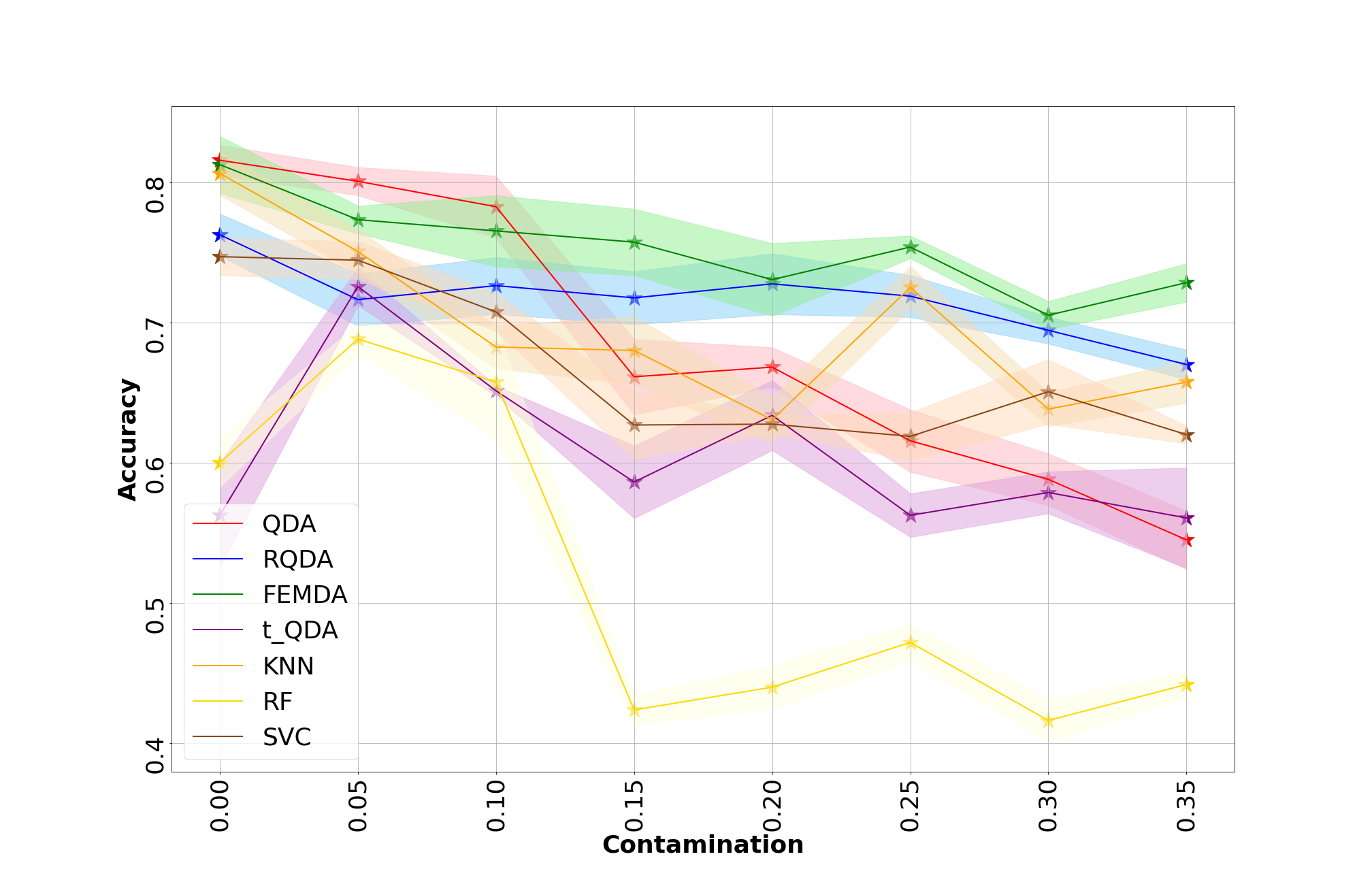}}
\caption{Accuracy evolution of the different algorithms when a percentage of the real train set is replaced by mild Gaussian noise}
\label{fig:8} 
\end{figure}

Similarly to simulated data on Avila and Frogs species, QDA is highly impacted by noise and quickly reaches a performance comparable to a random classifier. While having similar performances to FEMDA on clean datasets, RQDA appears to be significantly more impacted by contamination. On the Avila dataset, it almost reaches the performances of the non-robust QDA method. Although this unstructured noise does not significantly impact machine learning methods, FEMDA is again the method that handles the noise best, with the smallest decrease in performance. On smaller datasets, Glass and Sonar, QDA shows a better resilience to noise. Most methods aren't very affected by noise, and FEMDA emerges as the top-performing method when data is contaminated. While the authors \cite{HOUDOUIN2022Robust} showed that t-QDA and FEMDA achieved similar performances and robustness on both simulated and real datasets, t-QDA's efficiency drops quickly when the computational time is limited. To conclude, FEMDA proves to be a fast and robust method. FEMDA offers the best trade-off between computation time, robustness, and accuracy on real datasets and a real-time application framework. It can even compete with machine learning models when the dataset is small compared to the dimension.

\section{Conclusion and perspectives}
\label{sec:conc}

This chapter aims to provide readers with a tutorial presentation of discriminant analysis. In the second part, we introduce a general Bayesian framework that models the data of each cluster with an elliptically symmetric distribution. In this framework, each data point is allowed to have its own elliptically symmetric distribution with its own scale on the covariance matrix and a prior on the scale. The cluster's membership only imposes the mean vector and the scatter matrix. We demonstrate that we can retrieve the usual discriminant analysis methods by choosing the appropriate prior and density generator function. This chapter also provides a tractable way to deal with the general case using Jeffrey's prior, leading to the FEMDA algorithm. The third part is devoted to experiments. First, we conduct experiments on synthetic data, allowing us to evaluate various estimators' performances. Then, we focus on the classification performance for both simulated and real datasets. We gradually contaminate the data with uniform random noise to evaluate the robustness of each method. Experiments show that FEMDA outperforms the other discriminant analysis methods in most cases and can even compete with machine learning methods in some scenarios. FEMDA is also very resilient to outliers thanks to its robust mean and scatter matrix estimators, making it particularly suitable for handling noisy data. Faster than t-QDA, FEMDA is the statistical method with the best trade-off between performance, robustness, and computation time. 

In high dimensional regimes, Mahalanobis distance usually grows toward infinity, and t-QDA estimators and decision rules become very close to FEMDA. Future work aims to quantify the difference between these two methods in such a regime. 

\newpage

\bibliographystyle{spmpsci}
\bibliography{Semiparametric_Inference_in_ES}

\newpage

\appendix

\section{Proof of proposition 1.1}

According to the definition, the marginal likelihood of the sample $\mb{x}=(\mb{x}_1,...,\mb{x}_{n_k})$ belonging to cluster $\mathcal{C}_k$ is

\begin{align*}
    \mathcal{L}(\mb{x};\bs{\mu}_k, \bs{\Sigma}_k) &= \int_{]0,+\infty[^{n_k}} \prod_{i=1}^{n_k} p_{ik}\left(\mb{x}_i, t;\bs{\mu}_k, \bs{\Sigma}_k\right) dt. \\
    &= \int_{]0,+\infty[^{n_k}} \prod_{i=1}^{n_k} \frac{\Gamma(\frac{m}{2}) h_{i,k}(t)}{\pi^{\frac{m}{2}}\int_0^{+ \infty}u^{\frac{m}{2}-1}g_{i,k}(u)du}|\bs{\Sigma}_k|^{-\frac{1}{2}} t^{-\frac{m}{2}} \\
    & g_{i,k}\left( \frac{(\mb{x}_i-\bs{\mu}_k)^T \bs{\Sigma}_k^{-1} (\mb{x}_i-\bs{\mu}_k)}{t} \right) dt.
\end{align*}

Since we suppose that the observations are independent of each other, we have

\begin{align*}
    \mathcal{L}(\mb{x};\bs{\mu}_k, \bs{\Sigma}_k) =& \prod_{i=1}^{n_k} \frac{\Gamma(\frac{m}{2})|\bs{\Sigma}_k|^{-\frac{1}{2}} }{\pi^{\frac{m}{2}}\int_0^{+ \infty}u^{\frac{m}{2}-1}g_{i,k}(u)du} \int_0^{+\infty} h_{i,k}(t) t^{-\frac{m}{2}} \\
    & g_{i,k}\left( \frac{(\mb{x}_i-\bs{\mu}_k)^T \bs{\Sigma}_k^{-1} (\mb{x}_i-\bs{\mu}_k)}{t} \right)dt\\
    =& \prod_{i=1}^{n_k} \frac{\Gamma(\frac{m}{2})|\bs{\Sigma}_k|^{-\frac{1}{2}} }{\pi^{\frac{m}{2}}\int_0^{+\infty}u^{\frac{m}{2}-1}g_{i,k}(u)du} \\
    &\int_0^{\infty}h_{i,k}(t)t^{-\frac{m}{2}} g_{i,k}\left( \frac{(\mb{x}_i-\bs{\mu}_k)^T \bs{\Sigma}_k^{-1} (\mb{x}_i-\bs{\mu}_k)}{t} \right) dt.
\end{align*}

We then define $d_{i,k} = (\mb{x}_i-\bs{\mu}_k)^T \bs{\Sigma}_k^{-1} (\mb{x}_i-\bs{\mu}_k)$:

\begin{align*}
     \mathcal{L}(\mb{x};\bs{\mu}_k, \bs{\Sigma}_k) =& \prod_{i=1}^{n_k} \frac{\Gamma(\frac{m}{2})|\bs{\Sigma}_k|^{-\frac{1}{2}} d_{i,k}^{-\frac{m}{2}} }{\pi^{\frac{m}{2}} \int_0^{+\infty}u^{\frac{m}{2}-1}g_{i,k}(u)du} \\
     &\int_0^{\infty}h_{i,k}\left(d_{i,k}\frac{t}{d_{i,k}}\right) \left(\frac{d_{i,k}}{t}\right)^{\frac{m}{2}} g_{i,k}\left( \frac{d_{i,k}}{t} \right) dt \\
      =& \prod_{i=1}^{n_k} \frac{\Gamma(\frac{m}{2})|\bs{\Sigma}_k|^{-\frac{1}{2}} d_{i,k}^{1-\frac{m}{2}} }{\pi^{\frac{m}{2}}\int_0^{+\infty}u^{\frac{m}{2}-1}g_{i,k}(u)du} \\
      &\int_0^{\infty}\frac{d_{i,k}}{t^2} h_{i,k}\left(d_{i,k}\frac{t}{d_{i,k}}\right) \left(\frac{d_{i,k}}{t}\right)^{\frac{m}{2}-2} g_{i,k}\left( \frac{d_{i,k}}{t} \right) dt \\
      &= \prod_{i=1}^{n_k} \frac{\Gamma(\frac{m}{2})|\bs{\Sigma}_k|^{-\frac{1}{2}} d_{i,k}^{1-\frac{m}{2}} }{\pi^{\frac{m}{2}}\int_0^{+\infty}u^{\frac{m}{2}-1}g_{i,k}(u)du} \int_0^{\infty} h_{i,k}\left(\frac{d_{i,k}}{t}\right) t^{\frac{m}{2}-2} g_{i,k}(t) dt.
\end{align*}

\section{Proof of theorem 2.1}

We derive the estimators of location and scale by voiding the differential of the log-marginal likelihood $l$ by the parameter of interest. Assuming that $h_{i,k}$ is differentiable and that the integrals converge, one can write:

\begin{align*}
    \frac{\partial l}{\partial \bs{\mu}_k} =& \sum_{i=1}^{n_k} \left(1-\frac{m}{2}\right) \frac{2\bs{\Sigma}_k^{-1}(\mb{x}_i-\bs{\mu}_k)}{(\mb{x}_i-\bs{\mu}_k)^T \bs{\Sigma}_k^{-1} (\mb{x}_i-\bs{\mu}_k)} \\
    &+ \frac{\int_0^{\infty} \frac{2\bs{\Sigma}_k^{-1}(\mb{x}_i-\bs{\mu}_k)}{t} h_{i,k}'\left(\frac{(\mb{x}_i-\bs{\mu}_k)^T \bs{\Sigma}_k^{-1} (\mb{x}_i-\bs{\mu}_k)}{t}\right) t^{\frac{m}{2}-2} g_{i,k}(t) dt}{\int_0^{\infty} h_{i,k}\left(\frac{(\mb{x}_i-\bs{\mu}_k)^T \bs{\Sigma}_k^{-1} (\mb{x}_i-\bs{\mu}_k)}{t}\right) t^{\frac{m}{2}-2} g_{i,k}(t) dt} \\
    =& \bs{\Sigma}_k^{-1} \sum_{i=1}^{n_k} (\mb{x}_i-\bs{\mu}_k) \left[ \frac{2-m}{(\mb{x}_i-\bs{\mu}_k)^T \bs{\Sigma}_k^{-1} (\mb{x}_i-\bs{\mu}_k)} \right.\\
    &+ \left.\frac{2\int_0^{\infty} h_{i,k}'\left(\frac{(\mb{x}_i-\bs{\mu}_k)^T \bs{\Sigma}_k^{-1} (\mb{x}_i-\bs{\mu}_k)}{t}\right) t^{\frac{m}{2}-3} g_{i,k}(t) dt}{\int_0^{\infty} h_{i,k}\left(\frac{(\mb{x}_i-\bs{\mu}_k)^T \bs{\Sigma}_k^{-1} (\mb{x}_i-\bs{\mu}_k)}{t}\right) t^{\frac{m}{2}-2} g_{i,k}(t) dt} \right].
\end{align*}

Thus,

\begin{align*}
    \frac{\partial l}{\partial \bs{\mu}_k}(\hat{\bs{\mu}}_k) =& 0 \iff \hat{\bs{\mu}}_k = \frac{\sum_{i=1}^{n_k} w_{i,k} \mb{x}_i}{\sum_{i=1}^{n_k}w_{i,k}}, \mbox{with}\\ 
\end{align*}

\begin{align*}
    w_{i,k} =& \frac{1}{(\mb{x}_i-\bs{\hat{\mu}}_k)^T \bs{\hat{\Sigma}}_k^{-1} (\mb{x}_i-\bs{\hat{\mu}}_k)} \\
    &- \frac{2}{m-2} \frac{\int_0^{\infty} h_{i,k}'\left(\frac{(\mb{x}_i-\bs{\hat{\mu}}_k)^T \bs{\hat{\Sigma}}_k^{-1} (\mb{x}_i-\bs{\hat{\mu}}_k)}{t}\right) t^{\frac{m}{2}-3} g_{i,k}(t) dt}{\int_0^{\infty} h_{i,k}\left(\frac{(\mb{x}_i-\bs{\hat{\mu}}_k)^T \bs{\hat{\Sigma}}_k^{-1} (\mb{x}_i-\bs{\hat{\mu}}_k)}{t}\right) t^{\frac{m}{2}-2} g_{i,k}(t) dt} \\
    \frac{\partial l}{\partial \bs{\Sigma}_k^{-1}} =&  \frac{n_k}{2}\bs{\hat{\Sigma}}_k + \sum_{i=1}^{n_k} \frac{1-\frac{m}{2}}{(\mb{x}_i-\bs{\hat{\mu}}_k)^T \bs{\hat{\Sigma}}_k^{-1} (\mb{x}_i-\bs{\hat{\mu}}_k)}(\mb{x}_i-\bs{\hat{\mu}}_k)^T(\mb{x}_i-\bs{\hat{\mu}}_k)^T \\ 
    &+ (\mb{x}_i-\bs{\hat{\mu}}_k)(\mb{x}_i-\bs{\hat{\mu}}_k)^T \frac{\int_0^{\infty} h_{i,k}'\left(\frac{(\mb{x}_i-\bs{\hat{\mu}}_k)^T \bs{\hat{\Sigma}}_k^{-1} (\mb{x}_i-\bs{\hat{\mu}}_k)}{t}\right) t^{\frac{m}{2}-3} g_{i,k}(t) dt}{\int_0^{\infty} h_{i,k}\left(\frac{(\mb{x}_i-\bs{\hat{\mu}}_k)^T \bs{\hat{\Sigma}}_k^{-1} (\mb{x}_i-\bs{\hat{\mu}}_k)}{t}\right) t^{\frac{m}{2}-2} g_{i,k}(t) dt} \\
    =& \frac{n_k}{2}\left(\bs{\hat{\Sigma}}_k - \frac{m-2}{n_k} \sum_{i=1}^{n_k} w_{ik} (\mb{x}_i-\bs{\hat{\mu}}_k)(\mb{x}_i-\bs{\hat{\mu}}_k)^T\right).
\end{align*}

Similarly, 

\begin{align*}
    \frac{\partial l}{\partial \bs{\Sigma}_k}(\hat{\bs{\Sigma}}_k^{-1}) &= 0 \iff \hat{\bs{\Sigma}}_k = \frac{m-2}{n_k}\sum_{i=1}^{n_k} w_{i,k} (\mb{x}_i-\bs{\hat{\mu}}_k)(\mb{x}_i-\bs{\hat{\mu}}_k)^T.
\end{align*}

MMLE provides two coupled fixed-point equations that we solve iteratively, using the previous estimations $\hat{\bs{\mu}}_k^{(t)}$ and $\hat{\bs{\Sigma}}_k^{(t)}$ to estimate the weights at iteration $t+1$. Following Neyman-Pearson's lemma, the decision rule consists of choosing the cluster that maximizes the marginal likelihood computed using previously estimated parameters. The result is retrieved immediately after removing the multiplicative constants independent of $k$ in the likelihood obtained by Proposition 1.1.

\section{Proof of theorem 4}

Theorem 5 is simply the particular case of Theorem 2.1 with $h_{i,k}(t) \propto \frac{1}{t}$ and $h_{i,k}'(t) = -\frac{C}{t^2}$. Let us compute the new expression of $w_{i,k}$:

\begin{align*}
    w_{i,k} =& \frac{1}{(\mb{x}_i-\hat{\bs{\mu}}_k)^T\hat{\bs{\Sigma}}_k^{-1}(\mb{x}_i-\hat{\bs{\mu}}_k)}\\
    &-\frac{2}{m-2}\frac{\int_0^{+\infty}h_{i,k}'\left(\frac{(\mb{x}_i-\hat{\bs{\mu}}_k)^T\hat{\bs{\Sigma}}_k^{-1}(\mb{x}_i-\hat{\bs{\mu}}_k)}{t}\right)t^{\frac{m}{2}-3}g_{i,k}(t)dt}{\int_0^{+\infty}h_{i,k}\left(\frac{(\mb{x}_i-\hat{\bs{\mu}}_k)^T\hat{\bs{\Sigma}}_k^{-1}(\mb{x}_i-\hat{\bs{\mu}}_k)}{t}\right)t^{\frac{m}{2}-2}g_{i,k}(t)dt}
\end{align*}

\begin{align*}
    =& \frac{1}{(\mb{x}_i-\hat{\bs{\mu}}_k)^T\hat{\bs{\Sigma}}_k^{-1}(\mb{x}_i-\hat{\bs{\mu}}_k)}\\
    &+\frac{2}{m-2}\frac{\int_0^{+\infty}C\left(\frac{t}{(\mb{x}_i-\hat{\bs{\mu}}_k)^T\hat{\bs{\Sigma}}_k^{-1}(\mb{x}_i-\hat{\bs{\mu}}_k)}\right)^2t^{\frac{m}{2}-3}g_{i,k}(t)dt}{\int_0^{+\infty}C\frac{t}{(\mb{x}_i-\hat{\bs{\mu}}_k)^T\hat{\bs{\Sigma}}_k^{-1}(\mb{x}_i-\hat{\bs{\mu}}_k)}t^{\frac{m}{2}-2}g_{i,k}(t)dt} \\
    =& \frac{1}{(\mb{x}_i-\hat{\bs{\mu}}_k)^T\hat{\bs{\Sigma}}_k^{-1}(\mb{x}_i-\hat{\bs{\mu}}_k)}+\frac{2}{(m-2)(\mb{x}_i-\hat{\bs{\mu}}_k)^T\hat{\bs{\Sigma}}_k^{-1}(\mb{x}_i-\hat{\bs{\mu}}_k)}\\
    =& \frac{m}{m-2} \frac{1}{(\mb{x}_i-\hat{\bs{\mu}}_k)^T\hat{\bs{\Sigma}}_k^{-1}(\mb{x}_i-\hat{\bs{\mu}}_k)}.
\end{align*}

This allows us to retrieve the new expression of $\hat{\bs{\mu}}_k$ and $\hat{\bs{\Sigma}}_k$. Concerning the decision rule, we have 

\begin{align*}
    &\frac{|\hat{\bs{\Sigma}}_k|^{-\frac{1}{2}} \left( (\mb{x}_i-\hat{\bs{\mu}}_k)^T \hat{\bs{\Sigma}}_k^{-1}(\mb{x}_i-\hat{\bs{\mu}}_k)\right)^{1-\frac{m}{2}}}{\int_0^{+ \infty}t^{\frac{m}{2}-1}g_{i,k}(t)dt} \\
    &\int_0^{+\infty} h_{i,k}\left(\frac{(\mb{x}_i-\hat{\bs{\mu}}_k)^T \hat{\bs{\Sigma}}_k^{-1} (\mb{x}_i-\hat{\bs{\mu}}_k)}{t}\right) t^{\frac{m}{2}-2}g_{i,k}(t) dt \\
    &=\frac{C|\hat{\bs{\Sigma}}_k|^{-\frac{1}{2}} \left( (\mb{x}_i-\hat{\bs{\mu}}_k)^T \hat{\bs{\Sigma}}_k^{-1}(\mb{x}_i-\hat{\bs{\mu}}_k)\right)^{1-\frac{m}{2}}}{\int_0^{+ \infty}t^{\frac{m}{2}-1}g_{i,k}(t)dt} \\
    &\int_0^{+\infty}\frac{t}{(\mb{x}_i-\hat{\bs{\mu}}_k)^T \hat{\bs{\Sigma}}_k^{-1} (\mb{x}_i-\hat{\bs{\mu}}_k)} t^{\frac{m}{2}-2}g_{i,k}(t) dt \\
    =& C|\hat{\bs{\Sigma}}_k|^{-\frac{1}{2}} \left( (\mb{x}_i-\hat{\bs{\mu}}_k)^T \hat{\bs{\Sigma}}_k^{-1}(\mb{x}_i-\hat{\bs{\mu}}_k)\right)^{\frac{m}{2}}.
\end{align*}

Removing the constant $C$ that does not depend on $k$ and taking the logarithm allows us to retrieve the new expression of the decision rule.

\section{Retrieval of $t$-QDA} 

We will apply Theorem 2.1 and Remark 2.2 to retrieve the $t$-QDA method. First, we will derive the expression of the weight $w_{i,k}$. One has

\begin{itemize}
    \item $h_k(t) = \frac{\left(\frac{\nu_k}{2}\right)^{\frac{\nu_k}{2}}e^{-\frac{\nu_k}{2t}}}{\Gamma\left(\frac{\nu_k}{2}\right) t^{1+\frac{\nu_k}{2}}}$,
    \item $h_k'(t) =  \frac{\left(\frac{\nu_k}{2}\right)^{\frac{\nu_k}{2}}e^{-\frac{\nu_k}{2t}}}{2\Gamma\left(\frac{\nu_k}{2}\right) t^{3+\frac{\nu_k}{2}}} \left(\nu_k-\left(\nu_k+2\right)t\right)$.
\end{itemize}

We define $d_{i,k} = (\mb{x}_i-\hat{\bs{\mu}}_k)^T \hat{\bs{\Sigma}}_k^{-1}(\mb{x}_i-\hat{\bs{\mu}}_k)$. 

\begin{itemize}
    \item $h_k\left(\frac{d_{i,k}}{t}\right) = \frac{\left(\frac{\nu_k}{2}\right)^{\frac{\nu_k}{2}}e^{-\frac{\nu_kt}{2d_{i,k}}}}{\Gamma\left(\frac{\nu_k}{2}\right) d_{i,k}^{1+\frac{\nu_k}{2}}}t^{1+\frac{\nu_k}{2}}$,
    \item $h_k'\left(\frac{d_{i,k}}{t}\right) = \frac{\left(\frac{\nu_k}{2}\right)^{\frac{\nu_k}{2}}e^{-\frac{\nu_kt}{2d_{i,k}}}}{2\Gamma\left(\frac{\nu_k}{2}\right) d_{i,k}^{2+\frac{\nu_k}{2}}}t^{2+\frac{\nu_k}{2}} \left( \frac{\nu_k}{d_{i,k}}t - (\nu_k+2) \right)$.
\end{itemize}

We now focus on computing the ratio $\mathcal{I}$ of the two integrals in the expression of $w_{i,k}$:

\begin{align*}
    \mathcal{I} &= \frac{\int_0^{\infty}\frac{\left(\frac{\nu_k}{2}\right)^{\frac{\nu_k}{2}}e^{-\frac{\nu_kt}{2d_{i,k}}}}{2\Gamma\left(\frac{\nu_k}{2}\right) d_{i,k}^{2+\frac{\nu_k}{2}}}t^{2+\frac{\nu_k}{2}} \left( \frac{\nu_k}{d_{i,k}}t - (\nu_k+2) \right) t^{\frac{m}{2}-3}e^{-\frac{1}{2}t}dt}{\int_0^{\infty}\frac{\left(\frac{\nu_k}{2}\right)^{\frac{\nu_k}{2}}e^{-\frac{\nu_kt}{2d_{i,k}}}}{\Gamma\left(\frac{\nu_k}{2}\right) d_{i,k}^{1+\frac{\nu_k}{2}}}t^{1+\frac{\nu_k}{2}}t^{\frac{m}{2}-2}e^{-\frac{1}{2}t}dt} \\
    &= \frac{1}{2d_{i,k}} \frac{\int_0^{\infty}e^{-\frac{\nu_k+d_{i,k}}{2d_{i,k}}t}t^{\frac{\nu_k+m}{2}-1} \left( \frac{\nu_k}{d_{i,k}}t - (\nu_k+2) \right)dt}{\int_0^{\infty}e^{-\frac{\nu_k+d_{i,k}}{2d_{i,k}}t}t^{\frac{\nu_k+m}{2}-1}dt} \\
    &= \frac{1}{2d_{i,k}} \left( -(\nu_k+2) + \frac{\nu_k}{d_{i,k}}\frac{\int_0^{\infty}e^{-\frac{\nu_k+d_{i,k}}{2d_{i,k}}t}t^{\frac{\nu_k+m}{2}}dt}{\int_0^{\infty}e^{-\frac{\nu_k+d_{i,k}}{2d_{i,k}}t}t^{\frac{\nu_k+m}{2}-1}dt}\right) \\
    &= \frac{1}{2d_{i,k}} \left( -(\nu_k+2) + \nu_k\frac{\nu_k+m}{\nu_k+d_{i,k}}\right).
\end{align*}

Now we can obtain the expression of $w_{i,k}$ as follows:

\begin{align*}
    w_{i,k} &= \frac{1}{d_{i,k}} - \frac{2}{m-2} \mathcal{I} \\
    &= \frac{1}{d_{i,k}(m-2)} \left(m-2 + \nu_k + 2 - \nu_k\frac{\nu_k+m}{\nu_k+d_{i,k}} \right) \\
    &= \frac{\nu_k+m}{d_{i,k}(m-2)} \left( 1 - \frac{\nu_k}{\nu_k+d_{i,k}} \right) \\
    &= \frac{\nu_k+m}{(\nu_k+d_{i,k})(m-2)}.
\end{align*}

Now, we are going to use Remark 2.2 to obtain an estimator for $\nu_k$:

\begin{align*}
    \hat{\nu}_k =& \arg \max_{\nu \in \mathbb{R}^{+*}} \sum_{i=1}^{n_k} \log \left( \int_0^{\infty} \frac{\left(\frac{\nu}{2}\right)^{\frac{\nu}{2}}e^{-\frac{\nu t}{2d_{i,k}}}}{\Gamma\left(\frac{\nu}{2}\right) d_{i,k}^{1+\frac{\nu}{2}}}t^{1+\frac{\nu}{2}} t^{\frac{m}{2}-2}e^{-\frac{t}{2}}dt \right)\\
    =& \arg \max_{\nu \in \mathbb{R}^{+*}} \sum_{i=1}^{n_k} \log \left(\frac{1}{\Gamma\left(\frac{\nu}{2}\right)} \left(\frac{\nu}{2d_{i,k}}\right)^{\frac{\nu}{2}} \int_0^{\infty} e^{-\frac{\nu+d_{i,k}}{2d_{i,k}}t} t^{\frac{\nu+m}{2}-1}dt \right) \\
    =& \arg \max_{\nu \in \mathbb{R}^{+*}} \sum_{i=1}^{n_k} \log \left( \frac{\left(2d_{i,k}\right)^{\frac{m}{2}}\nu^{\frac{\nu}{2}}}{\Gamma\left(\frac{\nu}{2}\right) \left(\nu+d_{i,k}\right)^{\frac{\nu+m}{2}}} \right. \\
    &\left. \int_0^{\infty} \frac{\nu+d_{i,k}}{2d_{i,k}}e^{-\frac{\nu+d_{i,k}}{2d_{i,k}}t} \left(\frac{\nu+d_{i,k}}{2d_{i,k}}t\right)^{\frac{\nu+m}{2}-1}dt \right) \\
    =& \arg \min_{\nu \in \mathbb{R}^{+*}} \sum_{i=1}^{n_k} \log \left( \frac{\Gamma\left(\frac{\nu}{2}\right) \nu^{\frac{m}{2}} \left(1+\frac{d_{i,k}}{\nu}\right)^{\frac{\nu+m}{2}}}{\Gamma\left(\frac{\nu+m}{2}\right)} \right)\\
    =& \arg \min_{\nu \in \mathbb{R}^+} \frac{m}{2}\log(\nu) + \log \Gamma\left( \frac{\nu}{2} \right) - \log \Gamma\left( \frac{\nu + m}{2} \right) + \frac{\nu + m}{2n_k} \sum_{i=1}^{n_k} \log\left( 1 + \frac{d_{i,k}}{\nu} \right).
\end{align*}

Finally, we retrieve the $t$-QDA decision rule using Theorem 3:

\begin{align*}
     k =& \arg \max_{k \in [1,K]} \frac{|\hat{\bs{\Sigma}}_k|^{-\frac{1}{2}} d_{i,k}^{1-\frac{m}{2}}}{\int_0^{+ \infty}t^{\frac{m}{2}-1}e^{-\frac{t}{2}}dt} \int_0^{\infty} \frac{\left(\frac{\hat{\nu}_k}{2}\right)^{\frac{\hat{\nu}_k}{2}}e^{-\frac{\hat{\nu}_kt}{2d_{i,k}}}}{\Gamma\left(\frac{\hat{\nu}_k}{2}\right) d_{i,k}^{1+\frac{\hat{\nu}_k}{2}}}t^{1+\frac{\hat{\nu}_k}{2}} t^{\frac{m}{2}-2}e^{-\frac{t}{2}}dt \\
     =& \arg \max_{k \in [1,K]} \frac{|\hat{\bs{\Sigma}}_k|^{-\frac{1}{2}}\left(\frac{\hat{\nu}_k}{2}\right)^{\frac{\hat{\nu}_k}{2}}d_{i,k}^{-\frac{\hat{\nu}_k+m}{2}}}{\Gamma\left(\frac{\hat{\nu}_k}{2}\right)} \int_0^{\infty} e^{-\frac{\hat{\nu}_k+m}{2d_{i,k}}t}t^{\frac{\hat{\nu}_k+m}{2}-1}dt \\
     =& \arg \max_{k \in [1,K]} \frac{|\hat{\bs{\Sigma}}_k|^{-\frac{1}{2}}\left(\frac{\hat{\nu}_k}{2}\right)^{\frac{\hat{\nu}_k}{2}}d_{i,k}^{-\frac{\hat{\nu}_k+m}{2}}}{\Gamma\left(\frac{\hat{\nu}_k}{2}\right)} \left(\frac{2d_{i,k}}{\hat{\nu}_k+d_{i,k}}\right)^{\frac{\hat{\nu}_k+m}{2}} \Gamma\left( \frac{\hat{\nu}_k+m}{2} \right) \\
     =& \arg \max_{k \in [1,K]} \frac{\Gamma\left(\frac{\hat{\nu}_k+m}{2}\right) \hat{\nu}_k^{-\frac{m}{2}}}{\Gamma\left(\frac{\hat{\nu}_k}{2}\right)} |\bs{\Sigma}_k|^{-\frac{1}{2}} \left(1+\frac{d_{i,k}}{\hat{\nu}_k}\right)^{-\frac{\hat{\nu}_k+m}{2}} \\
 \end{align*}

 \begin{align*}
     =& \arg \min_{k \in [1,K]} \left(1+\frac{\hat{\nu}_k}{m}\right) \log \left(1 + \frac{(\mb{x}_i-\hat{\bs{\mu}}_k)^T \hat{\bs{\Sigma}}_k^{-1}(\mb{x}_i-\hat{\bs{\mu}}_k)}{\hat{\nu}_k} \right) \\
     &+ \log(\hat{\nu}_k) + \frac{1}{m} \log\left(|\hat{\bs{\Sigma}}_k|\right) + \frac{2}{m} \log \left( \frac{ \Gamma\left(\frac{\hat{\nu}_k}{2}\right)}{\Gamma\left(\frac{\hat{\nu}_k+m}{2}\right)} \right).
 \end{align*}
 
\end{document}